%% file: main.tex
\documentclass[10pt,twocolumn,letterpaper]{article}

\usepackage[pagenumbers]{cvpr} 

\input{preamble}

\definecolor{cvprblue}{rgb}{0.21,0.49,0.74}
\usepackage[pagebackref,breaklinks,colorlinks,allcolors=cvprblue]{hyperref}
\newcommand{\method}{Pegas}
\newcommand{\nothing}[1]{{}}

\def\confName{CVPR}
\def\confYear{2027}

\title{\method{}: Perceptually Guided Heuristics for Accelerating Diffusion-based Video Streaming}

\author{Henry Kam\textsuperscript{*}\quad
Bingxuan Li\textsuperscript{*}\quad
Colin Groth\quad
Finley Torrens\quad
Sai Qian Zhang\quad
Qi Sun\\
New York University\\
{\tt\small \{henrykam, bingxuan.li, cg4932, ft2377, sai.zhang, qisun\}@nyu.edu}\\[0.5em]
{\small\textsuperscript{*}Equal contribution.}
}

\begin{document}
\input{figures/tex/teaser}
\input{sec/0_abstract}

\input{sec/1_intro}
\input{sec/2_relatedworks}

\input{sec/3_method}
\input{sec/4_experiments}
\input{sec/5_limitations}

\input{sec/llm}
{
    \small
    \bibliographystyle{ieeenat_fullname}
    \bibliography{main}
}

\input{sec/X_suppl}

\end{document}

%% file: preamble.tex
\newcommand{\tablesize}{\fontsize{8}{9.5}\selectfont}
\providecommand{\x}{$\times$}

\usepackage{stfloats} 

%% file: figures/tex/teaser.tex
\twocolumn[{%
\renewcommand\twocolumn[1][]{#1}%
\maketitle
\begin{center}
\captionsetup{type=figure}
\includegraphics[width=\textwidth]{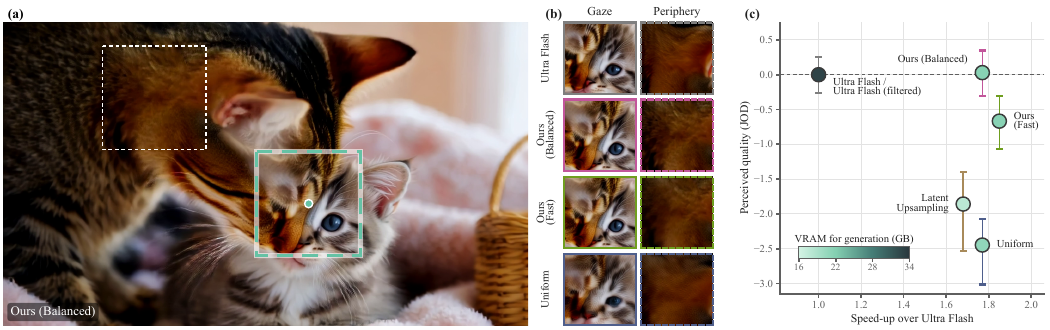}
\captionof{figure}{
\textbf{\method{} accelerates streaming video generation by concentrating computation near the viewer’s gaze.}
Across the pipeline (denoising, KV caching, super-resolution, and decoding), gaze guides how computation and memory are allocated to preserve detail near fixation.
(a) A frame generated with our method, with the viewer's $10^\circ$ fovea (dashed teal) and a peripheral region (dashed white).
(b) At the gaze, both configurations retain the fine detail around the kitten's eye seen in Ultra Flash. At the same throughput as \textit{Balanced}, Uniform softens this detail. The fur also shows the peripheral trade-off in \textit{Fast}.
(c) Perceived quality under central fixation versus throughput relative to Ultra Flash. Color indicates peak VRAM. \textit{Balanced} delivers $1.77\times$ the throughput and scores $+0.03$\,JOD relative to Ultra Flash, while Uniform at the same throughput scores 2.45\,JOD below Ultra Flash.
}
\label{fig:teaser}
\end{center}%
}]


%% file: sec/0_abstract.tex
\begin{abstract}
 Autoregressive video diffusion models enable streaming video generation but face substantial computational and memory costs that limit their efficiency and output resolution. Existing approaches commonly allocate resources uniformly across the visual field, overlooking the spatially varying sensitivity of human perception. We present \method{}, a perceptually guided heuristic for accelerating diffusion-based video streaming that allocates computation and memory throughout a frozen streaming video generator based on the viewer's gaze. \method{} integrates three key components: (1) gaze-guided generation that focuses late denoising steps near the gaze and caches past tokens according to perceptual importance; (2) foveated super-resolution that selectively refines gaze-centered regions while preserving global context through full-frame attention; (3) foveated decoding that reduces peripheral reconstruction costs through spatially adaptive resolution. To accommodate continuously changing gaze during streaming generation, we further introduce incremental super-resolution that dynamically updates refinement regions using the latest gaze measurements. 
Evaluated on an Ultra Flash backbone across 158 prompts with simulated human eye movements, \method{} achieves a 1.85$\times$ speed-up on one L40S and reduces mean peak memory from 32.8 to 21.7\,GB at 1K. In a separate user study under central fixation, viewers rated \textit{Balanced} level with Ultra Flash (+0.03 JOD) in overall quality and well above a uniform baseline at the same throughput.





\end{abstract}

%% file: sec/1_intro.tex
\section{Introduction}
\label{sec:intro}
Recent advances in autoregressive diffusion models enable new interactive visual experiences, where video is generated continuously in response to user input.  Unlike offline video generation, such applications require frames to be generated incrementally under strict latency constraints. However, generating high-resolution video at interactive frame rates remains computationally demanding, imposing substantial costs in both inference time and GPU memory. 
For example, Wan2.1~\cite{wan2025wan} takes 50 denoising steps and produces less than one frame per second even at 480p, and doubling the resolution quadruples the number of tokens processed at each step.
Existing approaches largely allocate computation uniformly across the visual field, despite the highly nonuniform spatial sensitivity of human vision: our visual acuity is highest at the foveal vision and decreases rapidly toward the periphery, suggesting that computation can be reduced in peripheral regions with limited perceptual impact. With eye tracking increasingly available in virtual and augmented reality (VR/AR) headsets, gaze measurements provide a practical way to allocate resources according to perceptual importance~\cite{patney2016towards}. This motivates a fundamental question:~\textit{can we accelerate streaming video generation by adapting computation and memory to human perception, without retraining existing generative models?}

Recent advances have substantially improved the efficiency of streaming video generation.
Distilled autoregressive generators produce video chunk by chunk in a few denoising steps~\cite{yin2025causvid,huang2025selfforcing}, while cascaded approaches such as Ultra Flash~\cite{huang2026ultraflashscalingrealtime} reach high resolution by super-resolving a low-resolution stream.
Nevertheless, high-resolution generation remains expensive, as denoising, super-resolution, decoding, and temporal attention continue to incur substantial computational and memory costs. 
These methods ultimately save computation \textit{uniformly} across the super-resolved (SR) frame.
Foveated rendering offers a complementary approach by concentrating computation in regions where viewers are most sensitive to visual detail. 
Prior work applies foveation to video reconstruction~\cite{kaplanyan2019deepfovea,wang2021focas} and mixed-resolution generation~\cite{chao2026foveated}.
However, existing foveated diffusion approaches rely on additional training to adapt generative models to multi-resolution representations. In contrast, we investigate \emph{training-free} foveation that preserves pretrained generative models while reducing computation and memory across the entire streaming pipeline.
\nothing{
We study its use in a frozen streaming cascade, where computation and historical memory must be allocated using gaze observed before the generated frames are displayed.
The generator must commit video 0.25 to 1.5\,s before it is shown, so the gaze has to be predicted online.
Moreover, every decision about where to spend quality is made before the system knows exactly where the viewer will look.
A streaming generator must also decide how much information from past frames to retain in its attention key-value (KV) cache, which grows as new frames are generated.
While keeping too much memory costs memory bandwidth, keeping too little causes the content to drift away from what the viewer has seen and may look at again.
Recent KV-cache methods ~\cite{zhang2023h2o,xiao2024streamingllm,li2024snapkv} choose what to keep by recency or by attention, never by where the viewer actually looks.
}

We introduce \method{}, \underline{Pe}rceptually \underline{G}uided Heuristics for \underline{A}ccelerating Diffusion-based Video \underline{S}treaming. Our key insight is that the spatially varying sensitivity of human vision can guide resource allocation throughout a \emph{pretrained} streaming generator, reducing computation and memory costs while preserving visual fidelity in perceptually important regions.
Specifically, \method{} integrates three main mechanisms.
First, in the generator, \emph{region-adaptive denoising} skips the late denoising steps away from the recent gaze, and a \emph{gaze-driven, multi-resolution KV cache} retains historical information at different spatial resolutions according to gaze, recency, and visual saliency. These components jointly reduce computation and memory without modifying the weights of the pretrained generative models. Second, \emph{foveated super-resolution} refines gaze-proximal regions while retaining full-frame attention to preserve global context. Finally, \emph{foveated decoding} reconstructs gaze-centered regions at full resolution and peripheral content at reduced resolution.
%
Notably, resource allocations must account for the delay between committing generated content and displaying it, which we call the \emph{commit horizon}.
Because the viewer's gaze can move during this interval, \method{} selects refinement regions using predictions derived from recent gaze measurements. It also runs super-resolution in short calls, allowing each call to incorporate the latest available gaze measurement.

We evaluate \method{} on 158 MovieGen Bench~\cite{polyak2024movie} prompts, using the same simulated gaze trajectory~\cite{kang2026infinitegaze} for all methods on each prompt. Within gaze-centered crops, we use KID~\cite{binkowski2018kid} to compare the distribution of generated content with that of Ultra Flash and MUSIQ~\cite{ke2021musiq} to assess local visual quality. 
To evaluate the video as a whole, we apply a gaze-dependent filter that accounts for the loss of peripheral acuity, then score the filtered clips with VBench~\cite{huang2024vbench} and HPSv2.1~\cite{wu2023human}. Additionally, we evaluate human preference in a controlled user study comparing our approach to alternative methods. 

On one L40S at 1K, \textit{Balanced} and \textit{Fast} deliver $1.77\times$ and $1.85\times$ the throughput of Ultra Flash with 34\% lower mean peak memory. At the throughput of \textit{Balanced}, its gaze-guided allocation scores 56.21 in gaze-crop MUSIQ versus 44.49 for Uniform. In a separate central-fixation study, the estimated quality difference between \textit{Balanced} and Ultra Flash was $+0.03$\,JOD. Both configurations fit on a 24\,GB desktop GPU at 1K and a single 48\,GB GPU at 2K.
In summary, our main contributions include:
\begin{itemize}
    \item A training-free framework for perceptually guided resource
    allocation in streaming video generation, accounting for the delay between generation and display.
    \item Gaze-guided generation that concentrates late denoising steps
    near the gaze and retains perceptually important regions at finer
    resolution in the KV cache.
    \item Gaze-contingent super-resolution and decoding that reduce computation devoted to peripheral content.
    \item A gaze-aware evaluation protocol that measures video generation quality at the gaze and across the full frame, complemented by a perceptual user study.
\end{itemize}

%% file: sec/2_relatedworks.tex
\section{Related Work}
\label{sec:related}

\paragraph{Streaming video generation.}
Video diffusion models such as Wan2.1~\cite{wan2025wan} jointly denoise clip frames over tens of steps. Autoregressive distillation enables streaming. CausVid~\cite{yin2025causvid} distills a bidirectional teacher into a few-step causal student, while Self Forcing~\cite{huang2025selfforcing} trains on the student's own rollouts using a KV cache.
High-resolution generation remains costly. FlashVSR~\cite{zhuang2025flashvsr} super-resolves streams in one diffusion step, while FlashDecoder~\cite{kang2026flashdecoder} accelerates decoding. Our backbone, Ultra Flash~\cite{huang2026ultraflashscalingrealtime}, combines a causal generator, latent upsampler, sparse one-step super-resolution transformer, and lightweight decoder for 1K and 2K output. These methods allocate computation independently of viewer gaze.

\vspace{-10pt}
\paragraph{Foveated rendering and gaze.}
Foveated rendering exploits declining visual acuity with eccentricity to reduce peripheral detail~\cite{guenter2012foveated,patney2016towards}. DeepFovea~\cite{kaplanyan2019deepfovea} learns to reconstruct the periphery from sparse samples. In video, foveated encoding reduces peripheral bitrate~\cite{illahi2020cloud,hsiao2021retina}, while foveated super-resolution restores detail near fixation~\cite{wang2021focas,lee2023crfp}.
These methods operate on existing scenes or video. A streaming generator must select regions hundreds of milliseconds before display, while future gaze remains unknown. This delay matters because tolerance for foveation depends on update latency, region size, and temporal stability~\cite{loschky2007late,albert2017latency}.
Approaches to gaze prediction include saliency models~\cite{droste2020unisal,linardos2021deepgaze}, generative gaze models~\cite{kang2026infinitegaze}, and saccade landing prediction~\cite{arabadzhiyska2017saccade}. \method{} selects regions from recent gaze measurements and uses a lightweight, separately trained saliency head only as a secondary cue for KV-cache retention.

\vspace{-10pt}
\paragraph{Adaptive computation and memory.}
Training-free diffusion methods merge similar tokens~\cite{bolya2023tome,bolya2023tomesd}, reuse predictions outside regions selected at each denoising step (RAS)~\cite{liu2025ras}, or denoise parts of the latent at lower resolution~\cite{jeong2025ralu,wu2025phasealigned}. Foveated Diffusion~\cite{chao2026foveated} adapts pretrained generators to a foveation mask that sets token resolution and decodes regions at different resolutions. Level-of-Token Diffusion~\cite{nakayama2026lot} extends this layout to variable-sized rectangular tokens. Both fix the layout before sampling. Streaming generators also select which past tokens to retain. Sliding-window KV caches keep recent frames~\cite{huang2025selfforcing}, while long-video methods retain early frames as attention sinks~\cite{yang2025longlive,liu2025rollingforcing}, following StreamingLLM~\cite{xiao2024streamingllm}. FramePack~\cite{zhang2025framepack} assigns fewer conditioning tokens to less important frames. Deep Forcing~\cite{yi2025deepforcing}, Dummy Forcing~\cite{guo2026dummyforcing}, and Forcing-KV~\cite{ji2026forcingkv} compress cached states by token or head. H2O~\cite{zhang2023h2o} and SnapKV~\cite{li2024snapkv} select entries using attention statistics. \method{} uses gaze available at chunk commitment to guide both denoising and cache allocation (\cref{sec:method-gaze}). Under a fixed token budget, gaze history, recency, and saliency determine which regions to retain in full, subsample, or drop, without extracting attention weights.

\vspace{-10pt}
\paragraph{Evaluating gaze-contingent generation.}
Foveated rendering metrics such as FovVideoVDP~\cite{mantiuk2021fovvideovdp} account for eccentricity by comparing outputs with full-quality references of the same content. Video generation benchmarks typically use whole-frame distributional and no-reference metrics~\cite{heusel2017fid,binkowski2018kid,huang2024vbench} without modeling how sensitivity varies with distance from gaze. Because our components can preserve or alter content, we measure reference-based fidelity for matched outputs and use distributional and no-reference metrics when content changes. We complement gaze-region scores with acuity-filtered whole-frame evaluation (\cref{sec:exp-metrics}).

%% file: sec/3_method.tex
\vspace{-1pt}
\section{Method}
\label{sec:method}
\begin{figure*}[t]
  \centering
  \includegraphics[width=\textwidth]{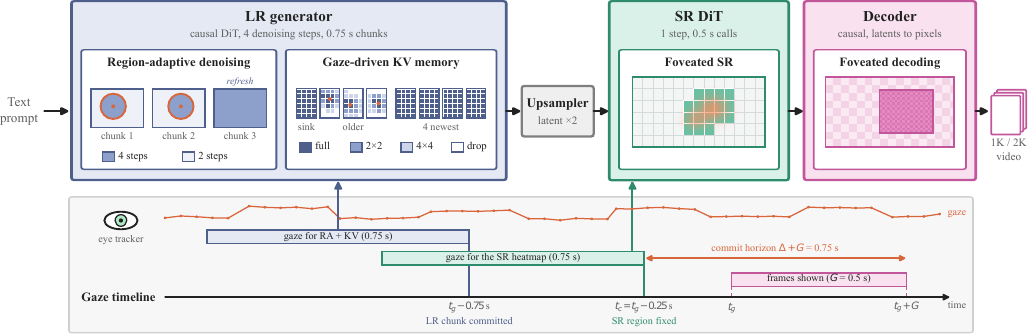}
  \caption{\textbf{\method{} on the Ultra Flash pipeline.}
  All four models stay frozen, but the resource allocation changes according to the viewer's gaze.
  The LR generator commits each 0.75\,s chunk 0.75\,s before it is shown, and the region-adaptive denoising and the gaze-driven KV memory use only the gaze measured in the 0.75\,s before that point (lower row).
  Each 0.5\,s SR call fixes its region 0.25\,s before its first frame is shown, and a heatmap of the preceding 0.75\,s of gaze selects the 22\% of SR blocks it refines, with any spare budget placed toward the screen center (see \cref{sec:method-heatmap}).
  The decoder decodes this SR window at full resolution and the rest of the frame at half resolution.}
  \label{fig:pipeline}
\vspace{-10pt}
\end{figure*}

\method{} keeps all four Ultra Flash stages frozen (\cref{fig:pipeline}). The generator uses recent gaze to guide late denoising and KV retention. After latent upsampling, newer gaze samples guide super-resolution. 
The decoder converts the latents to video, decoding the full frame at full resolution in \textit{Balanced} or the periphery at half resolution in \textit{Fast}. The timeline shows why SR can follow more recent gaze: it selects its refinement region closer to display time than the generator commits its chunk.



\vspace{-1pt}
\subsection{Backbone Pipeline}
\label{sec:method-backbone}

Our backbone, Ultra Flash~\cite{huang2026ultraflashscalingrealtime}, is a four-stage streaming pipeline built on Wan2.1~\cite{wan2025wan}. Its causal Wan2.1-1.3B generator, distilled with Self Forcing~\cite{huang2025selfforcing}, generates $480\times832$ video in chunks of three latent frames (0.75\,s). After four denoising steps at $t\in\{1000,937,833,625\}$, a clean pass updates the KV cache. Subsequent chunks attend to the most recent 21 latent frames, matching the training clip length.
A lightweight upsampler doubles the latent spatial resolution, and a super-resolution diffusion transformer (SR DiT) adds high-frequency detail in one denoising step. The SR DiT uses block-sparse attention over $2\times8\times8$ tokens (frames, rows, columns) and caches the preceding frame group for temporal consistency. A causal decoder then produces 1K ($960\times1664$) or 2K ($1440\times2496$) frames.
Ultra Flash allocates computation uniformly across the frame. At 1K on a single L40S, the generator, SR DiT, and decoder take 102, 88, and 22\,ms per output frame, respectively. Together they account for 95\% of the 222\,ms end-to-end average, motivating gaze-guided acceleration at all three stages.




\subsection{Region-Adaptive Denoising}
\label{sec:method-ra}

Early denoising steps establish layout and motion, while later steps refine detail~\cite{liao2026tunerdit}. Each step accounts for one of five forward passes per chunk and roughly a fifth of generation time. Since fine detail matters most near fixation, \method{} restricts the final two steps to tokens near the gaze.

The generator commits each chunk 0.75\,s before display (\cref{sec:method-gaze}). To accommodate gaze motion during this delay, we define $\mathcal{F}$ as the tokens within $10^\circ$ of any gaze sample in the 0.75\,s preceding commitment. In the final two steps, only tokens in $\mathcal{F}$ are re-noised and denoised, with attention over the full chunk. All other tokens retain their step-2 estimate. Writing $\hat{\mathbf{x}}_0^{(k)}$ for the clean estimate after step $k$ and $\mathbf{M}_{\mathcal{F}}$ for the binary gaze mask, the output is
\begin{equation}
  \hat{\mathbf{x}}_0 =
  \mathbf{M}_{\mathcal{F}} \odot \hat{\mathbf{x}}_0^{(4)}
  + (1-\mathbf{M}_{\mathcal{F}}) \odot \hat{\mathbf{x}}_0^{(2)}.
  \label{eq:ra}
\end{equation}
The clean pass caches every token, including those in the less refined periphery. To limit error accumulation, \method{} applies region-adaptive denoising to two of every three chunks and fully denoises the third. \cref{sec:exp-ablation} compares this with skipping the final two steps across all tokens.


\subsection{Gaze-Driven KV Memory}
\label{sec:method-kv}

Each chunk attends to the entire cache of up to 21 latent frames, seven times the chunk length. Attention cost and memory usage grow with the number of cached tokens. Ultra Flash retains all regions at full resolution, including those the viewer has never fixated. \method{} allocates a fixed budget of $B$ tokens (25\% of the cache in our experiments), prioritizing past and likely future gaze locations while retaining sparse context elsewhere for scene coherence. The supplementary illustrates the cache in one rollout.


\paragraph{Retention score.}
After each chunk, \method{} scores each cached $2\times2$ cell $c$ as
\begin{equation}
  r_c = \alpha(a_c)\bigl(w_{\mathrm{g}}G_c + w_{\mathrm{p}}P_c + w_{\mathrm{s}}S_c\bigr),
  \label{eq:kvscore}
\end{equation}
where $G_c$ and $P_c$ are the maximum foveal weights over two gaze windows: the 0.75\,s before chunk commitment and the interval when the cell's frame was displayed, respectively. The weight is one within $3^\circ$ of gaze and falls linearly to zero at $13^\circ$. $S_c$ denotes visual saliency, and $\alpha(a_c)=\exp(-a_c/12)$ downweights older cells, with age $a_c$ measured in latent frames. We set $w_{\mathrm{g}}=1$ and $w_{\mathrm{p}}=w_{\mathrm{s}}=0.5$.

The eye tracker supplies gaze samples, and cell age is known. A 0.48M-parameter head trained once to match UNISAL~\cite{droste2020unisal} predicts saliency from generator latents in 0.7\,ms per frame (Supplementary). This score requires no attention weights. Attention-based policies such as H2O and SnapKV~\cite{zhang2023h2o,li2024snapkv} require weights that fused kernels such as FlashAttention~\cite{dao2022flashattention} do not store and must therefore recompute them outside the kernel.


\paragraph{Allocating the budget.}
To preserve scene context across chunks, \method{} retains lower-scoring regions more sparsely. After each chunk, it sets $\lambda$ to retain $B$ tokens. Each $2\times2$ cell keeps all four tokens if $r_c \ge \lambda$ and one if $\lambda/2 \le r_c < \lambda$. A $4\times4$ block whose cells all score below $\lambda/2$ keeps one token if its highest cell score reaches $\lambda/4$, and none otherwise. The first frame serves as an attention sink~\cite{xiao2024streamingllm} and is always retained in full, as are the four newest latent frames.

For a token representing $m$ original tokens, \method{} adds $\log m$ to its attention logit~\cite{bolya2023tome}:
\begin{equation}
  \mathrm{Attn}(\mathbf{q}) =
  \mathrm{softmax}\!\left(
    \frac{\mathbf{q}\mathbf{K}^{\top}}{\sqrt{d}} + \log \mathbf{m}
  \right)\mathbf{V}.
  \label{eq:logbias}
\end{equation}
The representative thus receives the combined weight of $m$ identical tokens. The supplementary details selection and cache reorganization without slowing generation.

\paragraph{Keeping positions within the trained range.}
The generator uses temporal rotary position embeddings (RoPE)~\cite{su2021roformer} and was trained on clips of 21 latent frames (about 5\,s). 
Longer streams exceed this positional range and exhibit periodic color washout after about 5\,s under every cache policy we tested, including Ultra Flash's sliding window.
Following related position-reset strategies~\cite{liu2025rollingforcing,yesiltepe2025infinityrope}, \method{} reindexes cached frames consecutively within the trained range after each update.
This preserves frame order while compressing gaps between distant frames and eliminates the observed washout at no computational cost.

\subsection{The Commit Horizon}
\label{sec:method-gaze}
Conventional foveated rendering can redirect detail toward a new fixation on the next frame, often within tens of milliseconds of a saccade~\cite{loschky2007late,albert2017latency}. Streaming video generation makes this decision earlier: a stage chooses where to spend computation before it generates a group of frames, and that choice persists while the group is displayed. Consider a viewer who shifts gaze just after an SR call selects its refinement region. Even if the gaze measurement was exact, the viewer may already be looking elsewhere when the first frame appears. The challenge is therefore to select a region that will cover the viewer's gaze over a future interval.

We call the interval from a stage's gaze read to the end of its last frame's display the \emph{commit horizon}. For an SR call that reads gaze at $t_c$ and produces frames displayed over $[t_g,t_g+G)$, the horizon is $\Delta+G$, where $\Delta=t_g-t_c$. The two terms capture different sources of delay. The first, $\Delta$, is how old the gaze measurement is when the first frame appears. The second, $G$, is how long the same refinement decision remains in effect during playback. Even a stage that reads gaze shortly before display can therefore spend much of a long group refining a region the viewer has left.

This distinction motivates how we schedule super-resolution. Ultra Flash processes 1.5\,s of video in one SR call. At three to four saccades per second~\cite{john2021modelling}, a viewer can shift fixation several times while that group is on screen. We split it into three cache-linked 0.5\,s calls so that each can redirect refinement using newer gaze samples. 
In our offline evaluation, each SR call uses gaze samples recorded up to 0.25\,s before its first frame is scheduled to appear. Its eight frames then play at 16\,fps, so the same region remains fixed for another 0.5\,s. Together, the 0.25\,s lead and the 0.5\,s display interval give a 0.75\,s commit horizon.

\subsection{Predicting Where the Gaze Will Stay}
\label{sec:method-heatmap}

To keep the gaze within the refined region throughout the commit horizon, \method{} scores each latent token $i$ by its proximity to recent gaze samples. Using the display geometry to express token locations $\mathbf{x}_i$ and gaze samples $\mathbf{g}(\tau)$ in degrees of visual angle, we compute
\begin{equation}
  s_i = \sum_{\tau \in [t_c-T,\,t_c]}
  \exp\!\left(-\frac{\lVert \mathbf{x}_i-\mathbf{g}(\tau)\rVert^2}{2\sigma^2}\right),
  \label{eq:heatmap}
\end{equation}
where $T=0.75$\,s, $\sigma=2^\circ$, and $t_c$ precedes the first frame's display by 0.25\,s. We normalize the heatmap to a peak of one and add a weak central Gaussian prior (weight $10^{-3}$, width $12^\circ$) to rank blocks beyond recent gaze. Accounting for locations the viewer revisits improves gaze coverage over centering the region on current gaze (\cref{sec:exp-gaze}). The SR DiT refines the top $A=22\%$ of blocks, with $A$ controlling the trade-off between throughput and gaze coverage.

\subsection{Foveated Super-Resolution}
\label{sec:method-sr}

The upsampled latent captures the full frame, and the SR DiT primarily adds high-frequency detail. As sensitivity to fine detail falls with eccentricity, \method{} refines only blocks near the gaze and retains the upsampled latent elsewhere.

\paragraph{Full-frame context.}
Super-resolving a gaze-centered crop removes context at its boundary and reduces fidelity at the gaze (Supplementary). To preserve the full-frame context available during training, \method{} computes queries only for selected blocks $\mathcal B$ while retaining keys and values across the frame. With $\mathrm{KV}=(\mathbf K,\mathbf V)$ and $\Vert$ denoting concatenation along the token dimension,
\begin{equation}
\label{eq:fovsr}
\begin{aligned}
\mathbf Q &= \mathbf Q_{\mathcal B}, \\
\mathrm{KV} &= \mathrm{KV}_{\mathcal B}
  \mathbin{\Vert} \mathrm{KV}^{\mathrm{prev}}_{\bar{\mathcal B}}
  \mathbin{\Vert} \mathrm{KV}^{\mathrm{cache}}.
\end{aligned}
\end{equation}
Unselected blocks $\bar{\mathcal B}$ reuse their previous keys and values with temporal positions advanced by one group. As in Ultra Flash, $\mathrm{KV}^{\mathrm{cache}}$ supplies the preceding group's context. After each call, only selected blocks update their cache entries. We blend the refined region into the upsampled periphery over a $1^\circ$ band inside its boundary.

\subsection{Foveated Decoding}
\label{sec:method-dec}


After generation and super-resolution, the decoder still processes every pixel at full resolution, although the upsampled latent outside the SR region contains little high-frequency detail. The convolutional decoder supports both resolutions with the same weights, so our \textit{Fast} configuration decodes the refined region at full resolution and the periphery at half resolution. For each 1.5\,s group, \method{} takes the bounding box of its three 0.5\,s SR regions on the latent grid, pads it by four latent pixels for context, and rounds its boundaries outward to multiples of four. It decodes this window at full resolution. For the periphery, a second pass decodes the entire latent after factor-of-two average pooling and bilinearly upsamples the result. We blend the decoded window into this frame with a weight that tapers smoothly from one to zero over 32 output pixels along each window edge, except where it meets the frame boundary.

\paragraph{Preserving decoder memory.}
The causal decoder reads features from earlier frames. To preserve this history as the window moves, \method{} maintains a full-frame memory canvas at output resolution. After each call, it stores full-resolution features inside the window and upsampled features from the half-resolution pass elsewhere. A moving window thus inherits the previous frame's features at its new location. Groups without a new SR region, including the first, are decoded entirely at full resolution.

\input{tables/main_results}

\paragraph{Two operating points.}
We also offer a \textit{Balanced} configuration that decodes the whole frame at full resolution while retaining the same foveated SR, region-adaptive denoising and gaze-driven KV memory.
\textit{Fast} trades some peripheral sharpness for speed, and \textit{Balanced} keeps the periphery at full decoding resolution.

%% file: tables/main_results.tex
\begin{table*}[t]
  \centering
  \tablesize
  \setlength{\tabcolsep}{5pt}
    \begin{tabular}{@{}l cc cc ccccccc@{}}
    \toprule
    & \multicolumn{2}{c}{Cost} & \multicolumn{2}{c}{At-gaze quality} & \multicolumn{7}{c}{Acuity-filtered quality proxies ($\times100$)} \\
    \cmidrule(lr){2-3}\cmidrule(lr){4-5}\cmidrule(lr){6-12}
    Method
      & FPS (speed-up)$\uparrow$ & VRAM$\downarrow$
      & KID$\downarrow$ & MUSIQ$\uparrow$
      & SC$\uparrow$ & BC$\uparrow$ & MS$\uparrow$ & IQ$\uparrow$ & AQ$\uparrow$ & HPS$\uparrow$ & Avg.$\uparrow$ \\
    \midrule
    Ultra Flash~\cite{huang2026ultraflashscalingrealtime} & 4.50 (1.00\x) & 32.8 & 0.63 & 57.08 & 94.84 & 96.16 & 99.01 & 68.26 & 65.82 & 27.99 & 75.35 \\
    \midrule
    Foveated SR & 6.45 (1.43\x) & 25.4 & {3.07} & 55.93 & 94.45 & 95.61 & 99.03 & 68.10 & {65.05} & {27.63} & {74.98} \\
    \quad + Foveated Decode & 6.70 (1.49\x) & 25.4 & 3.22 & 55.73 & 93.80 & 95.41 & 99.04 & 64.74 & 64.27 & 26.97 & 74.04 \\
    \quad + Region-adaptive Denoising & 7.52 (1.67\x) & 25.4 & 4.16 & 55.95 & 93.75 & 95.40 & 99.03 & 64.50 & 64.30 & 26.94 & 73.99 \\
    \midrule
    \textbf{Ours (\textit{Balanced})} & 7.98 (1.77\x) & 21.7 & \textbf{4.38} & \textbf{56.21} & 94.00 & 95.44 & 99.06 & \textbf{68.19} & \textbf{64.93} & \textbf{27.57} & 74.86 \\
    \textbf{Ours (\textit{Fast})} & 8.31 (\textbf{1.85\x}) & 21.7 & 4.53 & 56.02 & 93.35 & 95.29 & \textbf{99.08} & 64.73 & 64.22 & 26.91 & 73.93 \\
    Latent Upsampling & 7.58 (1.68\x) & \textbf{18.0} & 50.19 & 50.87 & \textbf{94.54} & \textbf{96.47} & 99.04 & 67.95 & 64.40 & 27.47 & \textbf{74.98} \\
    Uniform (matched FPS) & 7.98 (1.77\x) & 21.0 & 59.88 & 44.49 & 94.05 & 95.34 & 98.99 & 58.97 & 62.81 & 25.99 & 72.69 \\
    \bottomrule
  \end{tabular}
  \caption{\textbf{Gaze-contingent acceleration on Ultra Flash at 1K.}
  At the same 7.98\,FPS, \textit{Balanced} improves gaze-crop MUSIQ~\cite{ke2021musiq} over Uniform by 11.72 points and also improves the acuity-filtered whole-frame average.
  Latent Upsampling reaches a similar whole-frame average to \textit{Balanced} but runs more slowly and scores lower at the gaze.
  \textit{Balanced} and \textit{Fast} reach $1.77\times$ and $1.85\times$ the throughput of Ultra Flash, respectively, with similar gaze-crop MUSIQ and a higher whole-frame average for \textit{Balanced}.
   Within the bottom four rows, \textbf{bold} values mark the best speed-up, lowest VRAM and best score in each quality column. 
  \emph{Avg.} is the unweighted mean of five VBench~\cite{huang2024vbench} scores and HPSv2.1~\cite{wu2023human} after gaze-dependent filtering at the $-1$\,JOD setting calibrated with FovVideoVDP~\cite{mantiuk2021fovvideovdp}.}
\vspace{-10pt}
  \label{tab:main}
\end{table*}

%% file: sec/4_experiments.tex
\providecommand{\pend}{{\color{red}\textbf{?}}}
\providecommand{\ours}{\textbf{Ours}}
\providecommand{\pilot}[1]{{\color{red}#1}}

\section{Experiments}
\label{sec:experiments}



\subsection{Experimental Setup}
\label{sec:exp-setup}

\paragraph{Backbone, test prompts, and gaze.}
We evaluate quality on all 158 MovieGen Bench~\cite{polyak2024movie} prompts released with Ultra Flash (\cref{tab:main,tab:others}). For each prompt, a video gaze model~\cite{kang2026infinitegaze} predicts a trajectory on the Ultra Flash output. Moving-gaze evaluations use this trajectory for all methods and a $47.9^\circ\times28.1^\circ$ field of view following GazeCom~\cite{dorr2010gazecom}. During generation, each stage allocates computation using only gaze samples available at decision time. We measure throughput on one L40S and compare GPUs in \cref{tab:gpus}.

\vspace{-8pt}
\paragraph{Video quality metrics.}
\label{sec:exp-metrics}
Even with the same prompt and seed, methods can generate different content, and the visibility of fine detail depends on distance from fixation. We therefore assess quality at the gaze and estimate whole-frame quality under gaze-dependent filtering.

For $10^\circ$ gaze-centered crops, we compute KID~\cite{binkowski2018kid} between each method and Ultra Flash references from four independent seeds using DINOv2-L features~\cite{oquab2024dinov2}. We assess local visual quality with the no-reference metric MUSIQ~\cite{ke2021musiq}.
For whole-frame evaluation, we use a gaze-dependent low-pass filter motivated by foveated rendering~\cite{guenter2012foveated,patney2016towards,kam2026dichoptic}. Its cutoff varies with eccentricity according to FovVideoVDP~\cite{mantiuk2021fovvideovdp}. We calibrated the filter on 36 Ultra Flash clips to a mean predicted loss of 1\,JOD with the original decoder. With the setting fixed, the mean predicted loss was 1.14\,JOD with the tiny decoder. We summarize the filtered outputs by an unweighted mean of five VBench dimensions~\cite{huang2024vbench} and framewise HPSv2.1~\cite{wu2023human}. This score describes quality under the assumed peripheral visibility model.

\input{tables/generator_compare}
\paragraph{User study.}
\label{sec:user_study}

To isolate the visible cost of peripheral approximation during steady fixation, we rendered all six conditions with gaze fixed at screen center. Participants were instructed to fixate a central cross and choose the higher-quality clip in a two-interval forced-choice trial. We used 20 prompts with sharp peripheral content and the same $47.9^\circ\times28.1^\circ$ field of view as the metrics. ASAP~\cite{mikhailiuk2020active} selected 80 pairs per participant. Of 11 observers, two met the leave-one-out exclusion criterion ($\mathrm{dist}_L>1.5$) set before data collection ended~\cite{perezortiz2017pwcmp}. We scaled the remaining 720 judgments to JOD relative to Ultra Flash and bootstrapped intervals over observers.
The supplementary gives the full protocol and analysis.

\vspace{-10pt}
\paragraph{Comparison methods.}
We first compare \textit{Balanced} and \textit{Fast} with other settings of the same Ultra Flash pipeline.
\textit{Ultra Flash} is the original pipeline at full quality.
\textit{Uniform} uses a sliding-window cache at the same 25\% token budget, skips the final two denoising steps across the frame, and runs full-frame SR at reduced spatial scale. In the reported L40S runs, its throughput matches \textit{Balanced} at 7.98\,FPS.
\textit{Latent Upsampling} omits the SR DiT and passes the upsampled latents directly to the full-resolution decoder.
We also include Ultra Flash after the acuity filter as a reference for ideal foveation, which removes detail only where the viewer can barely resolve it.
Finally, we compare our method with recent streaming generators, CausVid~\cite{yin2025causvid}, Causal Forcing~\cite{zhu2026causalforcing}, Self Forcing~\cite{huang2025selfforcing} and Dummy Forcing~\cite{guo2026dummyforcing}.

\input{figures/tex/qualitative_results}
\subsection{Main Results}
\label{sec:exp-main}

\paragraph{Quantitative results.}
At 1K on one L40S, \textit{Balanced} and \textit{Fast} reach $1.77\times$ and $1.85\times$ the throughput of Ultra Flash (\cref{tab:main}). At 7.98\,FPS, \textit{Balanced} scores 56.21 in gaze-crop MUSIQ versus 44.49 for Uniform. Its gaze-crop KID is also lower than Uniform's, although both accelerated methods depart from the Ultra Flash reference distribution. Latent Upsampling slightly exceeds \textit{Balanced} on the filtered whole-frame average (74.98 versus 74.86) while scoring lower at the gaze. The aggregate can therefore obscure a difference at fixation.

Under instructed central fixation, the estimated quality difference between \textit{Balanced} and Ultra Flash was $+0.03$\,JOD (95\% CI $[-0.31,0.35]$), with no significant difference detected (\cref{fig:teaser}c). Uniform scored $2.45$\,JOD below Ultra Flash at the same throughput as \textit{Balanced}. Thus, where computation was spent mattered even at matched speed. \textit{Fast} reached $1.85\times$ throughput and scored $-0.67$\,JOD relative to Ultra Flash, above Uniform. Including all 11 observers gave a similar estimate for \textit{Balanced} ($-0.13$\,JOD, $[-0.45,0.19]$). Filtered Ultra Flash, shown at the same marker as Ultra Flash, scored $+0.02$\,JOD ($[-0.27,0.25]$).
The supplementary material reports further analysis.


\paragraph{Comparison with other streaming generators.}
We compare the delivered outputs of four 480p generators with our 1K system, scoring the first 81 frames at native resolution (\cref{tab:others}). Under this protocol, \textit{Balanced} and \textit{Fast} rank first and second in gaze-crop MUSIQ. The ordering also holds on ten prompts after resizing all outputs to a common 1K display. \textit{Balanced} has the highest observed filtered average (75.36), close to Causal Forcing (75.20). Both configurations have higher reported throughput than the four external systems.

\paragraph{Qualitative comparison.}
\cref{fig:qualitative} complements the system-level comparison in \cref{tab:others} by showing the generated frames alongside crops at the gaze and in the periphery. Further visual results are shown in the supplementary.

\input{tables/Speed_results}

\subsection{Ablations}
\label{sec:exp-ablation}


\paragraph{Where does gaze guidance help?}
\label{sec:exp-content}

Shorter SR calls alone reduce peak memory from 32.8 to 27.9\,GB, and all five 25\% cache policies reach 24.2\,GB (\cref{tab:stages}). Gaze guides where to spend the reduced budget. Compared with Uniform Skip, region-adaptive denoising brings the gaze crop closer to Ultra Flash in PSNR, LPIPS and KID at a small throughput cost. At fixed cache capacity, Gaze KV has the highest PSNR and lowest LPIPS. Its PSNR lead is significant against every policy, its LPIPS lead over SnapKV and H2O is not, and paired KID comparisons do not separate the policies (\cref{tab:supp-stage-cis}). In a 36-prompt check with matched positional reindexing, Gaze KV improves PSNR over the sliding window by 1.29\,dB (95\% CI $[0.96,1.63]$).Together, these results show that gaze helps preserve detail where the viewer looks by guiding both refinement of the current frame and selection of context from earlier frames.

\paragraph{How well does gaze prediction work?}
\label{sec:exp-gaze}\label{sec:exp-oracle}
With a 0.25\,s lookahead, the refined region covered the viewer's gaze at display time in 84\% of samples across 158 prompts and 16 simulated viewers per prompt. Our predictor therefore concentrates detail where the viewer looks in most cases. To assess how closely this translates into oracle-level quality, we compared our method with an offline policy that uses the true gaze. Across 36 prompts, the oracle scored just 0.83 points higher on average in gaze-crop MUSIQ at similar throughput. Even with imperfect coverage, our predictor thus comes close to the oracle in visual quality where the viewer looks. The supplementary material details how the two policies place and update the refined region.


\paragraph{Throughput and memory across GPUs.}
The system comparison above measures throughput on a single L40S. Across a wider range of GPUs, the lower memory footprint of our method also expands the resolutions we can generate (\cref{tab:gpus}). At 2K on the H100, both configurations generate frames faster than Ultra Flash while reducing peak memory from 57.4\,GB to under 37\,GB. This brings the same 2K pipeline within the 48\,GB budget of the A6000 and L40S, where Ultra Flash runs out of memory. Both configurations also enable 1K streaming video generation on a single desktop GPU, the 24\,GB RTX 4090, where the original Ultra Flash pipeline exceeds device memory.

\input{tables/ablation1}

%% file: tables/generator_compare.tex
\begin{table*}[h]
  \centering
  \tablesize
  \setlength{\tabcolsep}{2.8pt}
  \resizebox{0.7\textwidth}{!}{
  \begin{tabular}{@{}l cc c ccccccc@{}}
    \toprule
    & \multicolumn{2}{c}{Cost} & At-gaze & \multicolumn{7}{c}{Acuity-filtered quality proxies ($\times100$)} \\
    \cmidrule(lr){2-3}\cmidrule(lr){4-4}\cmidrule(l){5-11}
    Method & FPS$\uparrow$ & VRAM$\downarrow$ & MUSIQ$\uparrow$
      & SC$\uparrow$ & BC$\uparrow$ & MS$\uparrow$ & IQ$\uparrow$ & AQ$\uparrow$ & HPS$\uparrow$ & Avg.$\uparrow$ \\
    \midrule
    CausVid~\cite{yin2025causvid} & 3.32 & 25.9 & 42.94 & 95.69 & 95.62 & 98.40 & 66.00 & 60.96 & 26.95 & 73.94 \\
    Causal Forcing~\cite{zhu2026causalforcing} & 6.55 & 25.1 & 41.86 & 95.98 & 95.64 & 98.52 & 69.91 & 63.02 & 28.12 & 75.20 \\
    Self Forcing~\cite{huang2025selfforcing} & 6.65 & 25.1 & 45.00 & \textbf{96.27} & 95.88 & 99.01 & 69.43 & 62.52 & 27.94 & 75.17 \\
    Dummy Forcing~\cite{guo2026dummyforcing} & 3.98 & \textbf{20.0} & 44.97 & 95.39 & 95.73 & 99.02 & \textbf{69.94} & 62.46 & \textbf{28.40} & 75.15 \\
    \textbf{Ours ({\textit{Balanced}})} & 7.98 & 21.7 & \textbf{55.75} & {95.49} & \textbf{96.26} & 99.01 & {68.03} & \textbf{65.41} & {27.94} & \textbf{75.36} \\
    \textbf{Ours (\textit{Fast})} & \textbf{8.31} & 21.7 & 55.68 & 94.96 & 96.12 & \textbf{99.03} & 65.41 & 64.88 & 27.41 & 74.63 \\
    \bottomrule
  \end{tabular}}
  
  \vspace{-5pt}
  \caption{\textbf{Speed and quality across streaming video generators.}
  Ours (\textit{Balanced}) achieves the highest acuity-filtered whole-frame average (\emph{Avg.}, as defined in \cref{tab:main}).
  \textit{Balanced} and \textit{Fast} rank first and second in gaze-crop MUSIQ.
  \textit{Fast} leads in throughput, followed by \textit{Balanced}, and both use less GPU memory than three of the four baselines.}
  \label{tab:others}
\vspace{-5pt}
\end{table*}

%% file: figures/tex/qualitative_results.tex
\begin{figure*}[t]
  \centering
  \includegraphics[width=\linewidth]{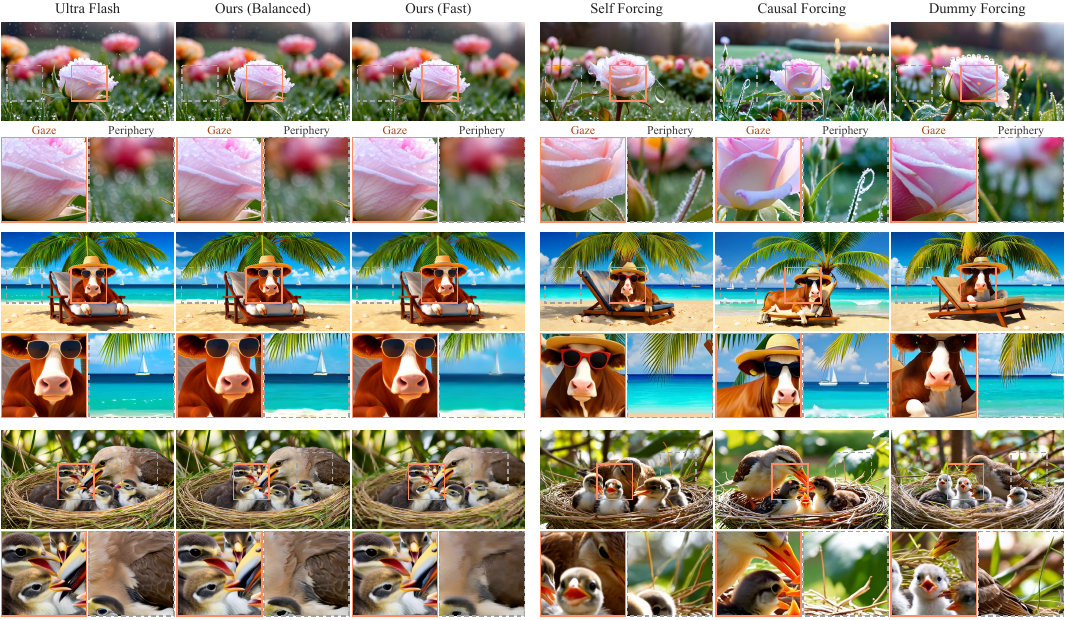}
  \caption{\textbf{Qualitative comparison at the gaze and in the periphery.} Solid orange and dashed gray boxes mark the gaze and peripheral crops, respectively. Both \textit{Balanced} and \textit{Fast} retain detail comparable to Ultra Flash at the gaze and generally show finer detail there than the external baselines. Peripheral detail is slightly softer in \textit{Balanced} and more visibly softened in \textit{Fast}.}
  \label{fig:qualitative}
\vspace{-10pt}
\end{figure*}


%% file: tables/Speed_results.tex
\begin{table}[t]
  \centering
  \tablesize
    \resizebox{0.49\textwidth}{!}{
  \setlength{\tabcolsep}{7pt}
  \begin{tabular}{@{}ll rr rr rr@{}}
    \toprule
    & & \multicolumn{2}{c}{\textit{Fast}} & \multicolumn{2}{c}{\textit{Balanced}} & \multicolumn{2}{c}{Ultra Flash} \\
    \cmidrule(lr){3-4}\cmidrule(lr){5-6}\cmidrule(l){7-8}
    GPU (VRAM) & Res. & FPS & GB & FPS & GB & FPS & GB \\
    \midrule
    RTX 4090 (24\,GB) & 1K & 7.04 & 21.9 & 6.83 & 21.9 & \multicolumn{2}{c}{OOM} \\
    A6000 (48\,GB) & 1K & 4.85 & 21.9 & 4.72 & 21.8 & 2.76 & 32.9 \\
     & 2K & 3.43 & 36.9 & 3.27 & 36.8 & \multicolumn{2}{c}{OOM} \\
    L40S (48\,GB) & 1K & 8.31 & 21.7 & 7.98 & 21.7 & 4.50 & 32.8 \\
     & 2K & 5.12 & 36.8 & 4.79 & 36.7 & \multicolumn{2}{c}{OOM} \\
    H100 (80\,GB) & 1K & 12.38 & 21.8 & 12.41 & 21.8 & 9.62 & 32.8 \\
     & 2K & 8.57 & 36.9 & 8.38 & 36.7 & 6.07 & 57.4 \\
    \bottomrule
  \end{tabular}}
  \caption{\textbf{Throughput and peak GPU memory across GPUs and resolutions.}
  \textit{Fast} and \textit{Balanced} both achieve higher throughput and lower peak memory than Ultra Flash in every configuration where all three fit on the GPU.
  Both also run at 1K on the 24\,GB RTX 4090 and at 2K on the 48\,GB A6000 and L40S, where Ultra Flash exceeds device memory (OOM).}
  \label{tab:gpus}
\vspace{-10pt}
\end{table}

%% file: tables/ablation1.tex
\begin{table}[t]
  \centering
  \tablesize
    \resizebox{0.49\textwidth}{!}{
  \setlength{\tabcolsep}{7pt}
  \begin{tabular}{@{}l cc ccc@{}}
    \toprule
    Variant & FPS (gain) & GB & PSNR$\uparrow$ & LPIPS$\downarrow$ & KID $\downarrow$ \\
    \midrule
    Ultra Flash & 4.42 (1.00\x) & 32.8 & ref & ref & 0.48 \\
    Ultra Flash (0.5\,s SR calls) & 4.49 (1.02\x) & 27.9 & 37.40 & 0.012 & 0.42 \\
    \midrule
    \multicolumn{6}{@{}l}{\emph{Fewer denoising steps}} \\
    Uniform Skip & 5.10 (1.16\x) & 27.9 & 18.96 & 0.251 & 2.11 \\
    RA (ours) & 4.86 (1.10\x) & 27.9 & \textbf{20.73} & \textbf{0.202} & \textbf{0.87} \\
    \midrule
    \multicolumn{6}{@{}l}{\emph{KV cache at 25\%}} \\
    Sliding window & 5.07 (1.15\x) & 24.2 & 19.55 & 0.277 & 2.61 \\
    StreamingLLM~\cite{xiao2024streamingllm} & 5.07 (1.15\x) & 24.2 & 19.80 & 0.275 & 2.70 \\
    SnapKV~\cite{li2024snapkv} & 4.99 (1.13\x) & 24.2 & 20.17 & 0.263 & 2.49 \\
    H2O~\cite{zhang2023h2o} & 4.97 (1.12\x) & 24.2 & 20.24 & 0.262 & \textbf{2.20} \\
    Gaze KV (ours) & 4.84 (1.10\x) & 24.2 & \textbf{20.50} & \textbf{0.260} & 2.60 \\
    \bottomrule
  \end{tabular}}
  \caption{\textbf{Ablation of gaze-guided denoising and KV retention.}
  The 0.5\,s Ultra Flash baseline splits each 1.5\,s super-resolution call into three shorter calls.
  At a small throughput cost over Uniform Skip, region-adaptive denoising more closely preserves the Ultra Flash output at gaze.
  With KV cache at 25\%, Gaze KV achieves the best gaze-crop PSNR/LPIPS among the cache policies; its PSNR lead is significant against every policy (\cref{tab:supp-stage-cis}).}
  \label{tab:stages}
\vspace{-10pt}
\end{table}

%% file: sec/5_limitations.tex
\section{Limitations and Future Work}
\label{sec:exp-limits}

We use simulated gaze~\cite{kang2026infinitegaze} as a flexible testbed and central fixation in the user study to ensure comparable viewing conditions. How well these results generalize to natural gaze during live viewing remains open.
Our sequential implementation reaches about 8\,FPS. Testing the full pipeline with live eye tracking at the target display rate requires higher throughput. Running pipeline stages concurrently across GPUs could shorten the commit horizon, a direction for future work.
Region-adaptive denoising and gaze-driven KV retention can alter content, while our evaluation focuses on visual quality. HPSv2.1, our only prompt-conditioned metric, yields comparable preference scores across methods but does not directly assess prompt alignment. Visual inspection revealed no clear mismatches, though systematic evaluation of prompt fidelity remains open.

\section{Conclusion}
We presented \method{}, training-free heuristics that allocate computation and memory in a pretrained streaming video framework according to human visual acuity.
On a single L40S, \method{} achieves up to $1.85\times$ the throughput of Ultra Flash at 1K with 34\% lower peak memory.  Under central fixation, viewers rated it close to Ultra Flash and well above Uniform at the same throughput.

%% file: sec/llm.tex
\section*{LLM Usage Statement}
We used LLMs under human supervision to edit the manuscript and code parts of the algorithm.

%% file: sec/X_suppl.tex
\clearpage
\setcounter{page}{1}
\maketitlesupplementary

\begingroup
\makeatletter
\patchcmd{\l@section}{1.5em}{1.4em}{}{\PackageError{pegas}{Could not widen contents numbering}{Check the section contents definition.}}
\def\pegas@tocprefix\hskip -1em.{\@ifnextchar~{\@gobble}{\pegas@tocprefix@nbs}}
\def\pegas@tocprefix@nbs\nobreakspace #1{}
\renewcommand{\numberline}[1]{\hb@xt@\@tempdima{#1.\hfil}\@ifnextchar\hskip{\pegas@tocprefix}{}}
\makeatother
\tableofcontents
\endgroup

%


\section{Comparisons at the same display resolution}
\label{sec:supp-display}
Table~2 in the main paper scores each generator at its native resolution, so the other methods are judged on their 832×480 output frames while ours are judged at 1664×960.
Viewers, however, see every method on the same resolution screen.
Enlarging them to 1K (which we assume as a common display resolution) lowers their gaze-crop MUSIQ and raises their filtered AQ (\cref{tab:supp-display}). At that common size, \textit{Balanced} and \textit{Fast} keep whole-frame quality on par with the other generators while running faster than all of them.
\begin{table*}[t]
  \centering
  \tablesize
  \begin{tabular}{@{}l c c ccccccc@{}}
    \toprule
    Method & FPS$\uparrow$ & MUSIQ$\uparrow$ & SC$\uparrow$ & BC$\uparrow$ & MS$\uparrow$ & IQ$\uparrow$ & AQ$\uparrow$ & HPS$\uparrow$ & Avg.$\uparrow$ \\
    \midrule
    CausVid & 3.32 & 38.74 & 97.21 & 97.06 & 98.80 & 63.43 & 60.22 & 27.23 & 73.99 \\
    Self Forcing & 6.65 & 38.71 & 97.44 & 97.34 & 99.28 & 66.50 & 63.03 & 28.29 & 75.31 \\
    Causal Forcing & 6.55 & 38.98 & 97.33 & 97.18 & 98.84 & 67.15 & 61.76 & 28.50 & 75.13 \\
    Dummy Forcing & 3.98 & 38.35 & 96.86 & 96.89 & 99.32 & 67.75 & 64.25 & 28.48 & 75.59 \\
    \midrule
    Ours (Balanced) & 7.98 & 57.30 & 97.39 & 96.62 & 99.25 & 68.04 & 63.77 & 28.73 & 75.63 \\
    Ours (Fast) & 8.31 & 57.24 & 96.80 & 96.31 & 99.25 & 65.83 & 63.47 & 28.24 & 74.98 \\
    \bottomrule
  \end{tabular}
  \caption{\textbf{Same quality at 1k resolution, with less compute.} Unlike Tab.~2 of the main paper, which scores each generator at its native resolution, this table shows every method at the same $1664\times960$ display size, as a viewer would see it: the $832\times480$ outputs of the other generators are enlarged bilinearly, and all methods are scored at that size on ten prompts of Tab.~2 of the main paper (first 81 frames). MUSIQ is on the gaze crop and the other columns are after the calibrated $-1$\,JOD filter ($\times100$, Avg.\ as in Tab.~1 of the main paper); FPS is each method's generation rate from Tab.~2 of the main paper. \textit{Balanced} has the highest filtered average and \textit{Fast} is within 0.61 of the best other generator, while both generate more frames per second than any of them.}
  \label{tab:supp-display}
\end{table*}

\paragraph{Timing protocol and baselines.}
\label{sec:supp-timing-protocol}
We time \method{} and Ultra Flash on one L40S over the 36 timing prompts of Tab.~1 of the main paper, generating 165 frames at $1664\times960$ for 16\,fps playback with the tiny decoder (TAEHV), \texttt{torch.compile} disabled and one latent frame decoded at a time. FPS covers the 165 frames, and VRAM is the mean peak allocated memory over the 36 prompts.
The external systems of Tab.~2 of the main paper are timed after warm-up with cached text embeddings, decoding included and video writing disabled, as 81 frames divided by the time to deliver the scored prefix, including 26\,ms for the 480p display resize, with the maximum peak allocated VRAM over six prompts. Self Forcing and Causal Forcing generate 81 frames, Dummy Forcing generates 165 frames and decodes once, CausVid needs two overlapping rollouts to reach frame 81, and FlashVSR produces 93 frames at its first eligible output boundary.
The other rows of Tab.~3 of the main paper give mean FPS and the largest per-prompt peak allocation over six prompts at 1K or three at 2K ($2496\times1440$), and OOM includes the 2K A6000 run that fell back to host RAM.
Each clip is seeded with its prompt index and the LR generator's random state is fixed, so comparisons that vary only SR or decoding share the same LR video. Generation is bit-reproducible on one GPU type but not across types, so full-reference comparisons pair same-type clips, and the per-prompt values of Tab.~1 of the main paper and of the cross-method rows (L40S clips) are not paired with those of Tab.~4 of the main paper (H200 clips), although distribution metrics such as KID remain comparable.
Uniform keeps a 25\% sliding-window cache with a sink frame, skips the final two denoising steps for every token on the two-of-three chunk schedule of RA, and runs the upsampler, SR and decoder at 0.75 of the linear output size, chosen in a short search over scales, before bicubic resizing to $1664\times960$. It reaches 7.98\,FPS (range 7.82--8.10) in timing-only runs, the same as \textit{Balanced}.
Latent Upsampling keeps Ultra Flash's full generator but drops the SR DiT and decodes the upsampled latents at full resolution. Viewers placed it 1.89\,JOD below \textit{Balanced}, so the advantage of gaze allocation does not depend on how Uniform is built.

\section{Our region policy on FlashVSR}
\label{sec:supp-flashvsr}
Beyond comparing \method{} with other systems (\cref{sec:supp-display}), we can apply its heuristics to them: they need no training, so they are not tied to Ultra Flash. To test this, we apply our SR region policy (recent-gaze heatmap, 22\% of blocks fixed 0.25\,s ahead, queries only in the region) to FlashVSR~\cite{zhuang2025flashvsr}, a different streaming super-resolution model, without retraining it (\cref{tab:supp-flashvsr}).
The policy makes FlashVSR 1.49\x{} faster, and 2.02\x{} faster when its decoder also runs at full resolution only inside the region, while keeping about 76\% of FlashVSR's gain in gaze-crop MUSIQ over no SR. Against full FlashVSR on 158 prompts with viewer~0's gaze, the gaze lies inside the region in 87.7\% of frames, and PSNR inside the region is 31.1\,dB, against 26.4\,dB for bicubic upsampling of the input; the decode window changes quality by at most 0.02.
Foveated FlashVSR adds no flicker: VBench temporal flickering (higher means less flicker) changes by $+0.01$ after the viewing filter and by $+0.07$ on raw frames.

\begin{table}[t]
  \centering
  \tablesize
  \setlength{\tabcolsep}{3pt}
  \resizebox{\columnwidth}{!}{%
  \begin{tabular}{@{}l cc c c@{}}
    \toprule
    & \multicolumn{2}{c}{FPS (speed-up)$\uparrow$} & VRAM & Gaze-crop MUSIQ \\
    \cmidrule(lr){2-3}
    FlashVSR & End to end & FlashVSR alone & (GB)$\downarrow$ & change$\uparrow$ \\
    \midrule
    Full & 3.38 & 7.68 & 29.8 & -- \\
    Foveated & 3.95 (1.17\x{}) & 11.44 (1.49\x{}) & 29.8 & $-1.79$ \\
    \quad + decode window & 4.34 (\textbf{1.29\x{}}) & 15.54 (\textbf{2.02\x{}}) & \textbf{29.2} & \textbf{\boldmath$-1.78$} \\
    No SR (480p) & -- & -- & -- & $-7.45$ \\
    \bottomrule
  \end{tabular}}
  \caption{\textbf{Foveated FlashVSR}: our region policy applied to FlashVSR without retraining, upsampling the $480\times832$ Self Forcing video by two to $960\times1664$. The decode window also runs FlashVSR's decoder at full resolution only inside the region. FPS and VRAM on the 36 timing prompts of Tab.~1 of the main paper, one L40S, with speed-ups against full FlashVSR. The MUSIQ change is on the gaze crop of simulated viewer~0 over 158 prompts, paired against full FlashVSR, which shares the LR video. Foveation makes FlashVSR alone 2.02\x{} faster and keeps about 76\% of its MUSIQ gain over no SR. Bold marks the best value in each column, excluding full FlashVSR and no SR.}
  \label{tab:supp-flashvsr}
\end{table}

\section{Speed against quality, per metric}
\label{sec:supp-speed-quality}
\Cref{fig:speed-quality} shows the speed--quality trade-off of Fig.~1c of the main paper and \cref{fig:user-study} in gaze-crop MUSIQ instead of the user-study JOD, for every row of Tab.~1 of the main paper including the component rows.
\input{figures/tex/speedVsQuality}

\section{Where the savings come from}
\label{sec:supp-attribution}
This section attributes the memory and time that \method{} saves to the components of the pipeline and checks that the gain from gaze holds across budgets.

\paragraph{Memory.}
Peak memory falls with the budget sizes, whichever tokens fill them (\cref{tab:supp-memory}). Every 25\% cache policy in Tab.~4 of the main paper peaks at 24.2\,GB, and Uniform (21.0\,GB) and Latent Upsampling (18.0\,GB) also fit 1K on a 24\,GB GPU. Gaze lets the budgets be this small without a perceived loss, since at a similar footprint Uniform scores 2.45\,JOD below Ultra Flash while \textit{Balanced} is not distinguished from it.

\begin{table}[t]
  \centering
  \tablesize
  \begin{tabular}{@{}lcc@{}}
    \toprule
    Configuration (cumulative) & Peak VRAM (GB) & Saving (GB) \\
    \midrule
    Ultra Flash & 32.8 & -- \\
    + SR in 0.5\,s calls (no gaze) & 27.9 & 4.9 \\
    + SR queries on 22\% of blocks & 25.4 & 2.5 \\
    + KV cache at 25\% (\textit{Balanced}) & 21.7 & 3.7 \\
    \bottomrule
  \end{tabular}
  \caption{\textbf{Peak GPU memory as each change is added to Ultra Flash}, on one L40S at 1K (lower is better), from Tabs.~1 and~4 of the main paper. Each row adds one change to the row above, and the saving is relative to that row. Together they bring \textit{Balanced} from 32.8 to 21.7\,GB, within a 24\,GB GPU. Region-adaptive denoising does not change peak memory.}
  \label{tab:supp-memory}
\end{table}

\paragraph{Throughput.}
\Cref{tab:supp-profile} divides the time of each pipeline component by the frames it produces. On the L40S, foveated SR gives 69\% of the time saved and RA with the KV budget 31\%. Foveated decoding saves a further 5\% but raises the speed-up of FlashVSR alone from 1.49\x{} to 2.02\x{} (\cref{tab:supp-flashvsr}), so we recommend \textit{Balanced} by default and \textit{Fast} for pipelines with a costly decoder.
On the H100 the saving falls to 22\% and \textit{Fast} gains little over \textit{Balanced}, because the components \method{} shrinks are already cheap there. The savings are therefore largest where the generator's attention and the SR DiT dominate, as on the L40S and the A6000 (1.71 to 1.85\x{} at 1K, Tab.~3 of the main paper).
Summed component times give steady-state rates 1 to 3\% above end to end for Ultra Flash and 4.5 to 6\% above for ours, so steady-state L40S speed-ups would be 1.84\x{} and 1.94\x{} for \textit{Balanced} and \textit{Fast}, against 1.77\x{} and 1.85\x{}. We report end-to-end rates and use component times only to attribute the savings.

\begin{table}[t]
  \centering
  \tablesize
  \setlength{\tabcolsep}{2.5pt}
  \resizebox{\columnwidth}{!}{%
  \begin{tabular}{@{}ll rrrr r rr@{}}
    \toprule
    & & \multicolumn{5}{c}{Time per output frame (ms)$\downarrow$} & \multicolumn{2}{c}{FPS$\uparrow$} \\
    \cmidrule(lr){3-7}\cmidrule(l){8-9}
    GPU & Method & LR generator & SR DiT & Decoder & Other SR & Total & Steady state & End to end \\
    \midrule
    L40S & Ultra Flash & 102.9 & 86.8 & 21.8 & 8.6 & 220.1 & 4.54 & 4.50 \\
    & \textit{Balanced} & \textbf{71.7} & \textbf{20.3} & 21.9 & 6.0 & 119.9 & 8.34 & 7.98 \\
    & \textit{Fast} & 71.9 & \textbf{20.3} & \textbf{17.2} & \textbf{4.4} & \textbf{113.8} & \textbf{8.79} & \textbf{8.31} \\
    \midrule
    H100 & Ultra Flash & 54.4 & 29.8 & 15.8 & 2.5 & 102.5 & 9.76 & 9.49 \\
    & \textit{Balanced} & 49.6 & 13.0 & 15.8 & 2.1 & 80.5 & 12.43 & 11.86 \\
    & \textit{Fast} & \textbf{49.3} & \textbf{12.8} & \textbf{14.7} & \textbf{1.6} & \textbf{78.4} & \textbf{12.76} & \textbf{12.17} \\
    \bottomrule
  \end{tabular}}
  \caption{\textbf{Time per output frame by pipeline component} at 1K on one L40S and one H100, over the 36 timing prompts of Tab.~1 of the main paper with the tiny decoder. Each LR generator chunk is divided by its 12 output frames and each SR group by its 24, after the first chunk or group. Other SR is upsampling and overhead. Steady-state FPS inverts the total, and end-to-end FPS covers the complete 165-frame run, including startup. H100 rates differ slightly from Tab.~3 of the main paper, which averages six prompts. Most of the saving comes from the LR generator and the SR DiT, and \textit{Fast} also shortens the decoder. Bold marks the best value in each column within each GPU, excluding Ultra Flash.}
  \label{tab:supp-profile}
\end{table}

\begin{figure*}[t]
  \centering
  \includegraphics[width=\textwidth]{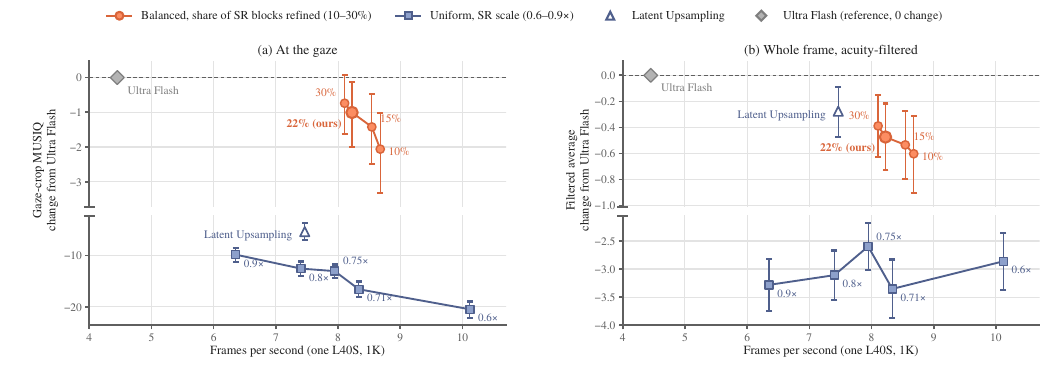}
  \caption{\textbf{Spending SR at the gaze keeps quality at every speed} (36 prompts, one L40S, 1K, tiny decoder). Each point is one configuration: its frame rate and its quality change from the Ultra Flash clip of the same job (0 is Ultra Flash; higher is better), with 95\% prompt-bootstrap intervals. \textit{Balanced} varies the share of SR blocks it refines (10 to 30\%); Uniform varies the scale of its whole-frame SR (0.6 to 0.9$\times$). Each y axis is broken between the configurations near Ultra Flash (top) and those far below it (bottom). Between 7.9 and 8.7\,FPS, \textit{Balanced} loses 0.7 to 2.1 MUSIQ points at the gaze and Uniform 13.0 to 16.6; over the whole frame, \textit{Balanced} loses 0.4 to 0.6 points and Uniform 2.6 to 3.4. Shrinking \textit{Balanced}'s budget from 30 to 10\% gains only about 0.6\,FPS.}
  \label{fig:supp-frontier}
\end{figure*}

\paragraph{Any budget.}
\Cref{fig:supp-frontier} varies the SR block budget of \textit{Balanced} (10 to 30\%) and the SR scale of Uniform (0.6 to 0.9) to test whether spending the savings at the gaze beats spending them uniformly at every speed, not only at the operating point of Tab.~1 of the main paper (the KV budget is varied in \cref{sec:supp-kv-budget}).
\textit{Balanced} stays within 2.1 gaze-crop MUSIQ points and 0.60 filtered-average points of Ultra Flash at 8.1 to 8.7\,FPS, whereas Uniform loses 9.8 to 20.5 and 2.6 to 3.4 points and raises gaze-crop KID ($\times10^3$) by 63 to 100 at every scale, even at 6.35\,FPS (scale 0.9).
Only the 10\% budget raises the KID of \textit{Balanced} clearly, by 11.3 (95\% CI [4.1, 19.1]), and Uniform has its best KID and filtered imaging quality at 0.75, the scale of Tab.~1 of the main paper.

\paragraph{Target hardware.}
\method{} targets a single workstation or consumer GPU, where Ultra Flash does not fit at 1K on a 24\,GB RTX 4090 (Tab.~3 of the main paper). Its training-free heuristics change only where each component spends its budget, so the speed-up is relative to the backbone, and they apply without retraining to faster backbones and hardware and to FlashVSR (\cref{sec:supp-flashvsr}).

\section{Reading the quality metrics}
\label{sec:supp-metric-reading}
Each automated metric sees only part of what viewers judge, so we relate the metrics of Tab.~1 of the main paper to one another and to the user study.

\paragraph{Component rows.}
In Tab.~1 of the main paper, foveated decoding lowers the filtered average from 74.98 to 74.04, mostly through imaging quality, while RA and the KV budget together cost 0.12 points for 0.34\x{} of throughput (\textit{Balanced} against foveated SR). Viewers also noticed foveated decoding, placing \textit{Fast} 0.67\,JOD below Ultra Flash and \textit{Balanced} 0.03\,JOD above it.

\paragraph{Agreement with the user study.}
Gaze-crop KID and MUSIQ in Tab.~1 of the main paper rank the four accelerated study conditions as viewers did. The filtered average instead places Latent Upsampling, which keeps Ultra Flash's full generator and leads on SC and BC, level with \textit{Balanced}, although viewers placed it 1.89\,JOD lower.
The whole-frame proxies thus check that the periphery and content stay intact, and the at-gaze metrics track the detail viewers judged at fixation. The comparison is indirect, as the study used central gaze.

Scoring the 120 study clips themselves shows which metrics agree with viewers (\cref{tab:study-metrics}). MUSIQ on a $2.5^\circ$ crop places filtered Ultra Flash level with Ultra Flash ($-0.04$, 95\% CI [$-0.10$, $0.02$]) and Latent Upsampling below it ($-2.35$, [$-3.29$, $-1.52$]), and only KID on a $20^\circ$ crop separates \textit{Fast} (25.0, [11.5, 44.3]) from \textit{Balanced} (11.5, [4.8, 21.7]), both as viewers did. Larger MUSIQ crops penalize the filter, which viewers did not see, but barely change \textit{Balanced} and \textit{Fast}, so the $10^\circ$ crop of Tab.~1 of the main paper, which has no filtered condition, is unaffected.
FovVideoVDP predicts more loss than viewers reported where content is shared, and for \textit{Balanced} and \textit{Fast} its scores (4.75 and 4.59) measure the content change from RA and the KV budget rather than a visible loss.
The crop sizes were chosen after seeing these data, and six conditions are few, so the agreement is consistent with the study rather than a validated metric.
\begin{table*}[t]
  \centering
  \tablesize
  \setlength{\tabcolsep}{6pt}
  \begin{tabular}{@{}l c ccc cc c@{}}
    \toprule
    & & \multicolumn{3}{c}{MUSIQ change$\uparrow$} & \multicolumn{2}{c}{KID ($\times10^3$)$\downarrow$} & \\
    \cmidrule(lr){3-5}\cmidrule(lr){6-7}
    Condition & User study (JOD)$\uparrow$ & $2.5^\circ$ crop & $5^\circ$ crop & $10^\circ$ crop & $5^\circ$ crop & $20^\circ$ crop & FovVideoVDP loss (JOD)$\downarrow$ \\
    \midrule
    Filtered Ultra Flash & $+0.02$ & $-0.04$ & $-1.00$ & $-5.31$ & 1.6 & 1.6 & 1.35 \\
    \textit{Balanced} & $+0.03$ & $-0.16$ & $-0.77$ & $-0.88$ & 2.5 & 11.5 & -- \\
    \textit{Fast} & $-0.67$ & $-0.21$ & $-0.79$ & $-0.87$ & 2.9 & 25.0 & -- \\
    Latent Upsampling & $-1.86$ & $-2.35$ & $-3.99$ & $-4.98$ & 70.2 & 41.6 & 2.35 \\
    Uniform & $-2.45$ & $-2.60$ & $-8.98$ & $-13.14$ & 52.8 & 38.2 & -- \\
    \midrule
    Rank correlation $\rho$ with the user study & -- & 0.77 & -- & 0.37 & -- & 0.71 & -- \\
    \bottomrule
  \end{tabular}
  \caption{\textbf{Metrics against the user study} on the 120 study clips, as condition means over 20 prompts (central gaze crop, the study's field of 47.9\,deg across at 34.7 pixels per degree), since each per-prompt JOD rests on only two to three trials per condition pair. User-study JOD and MUSIQ are differences from Ultra Flash, and KID (the estimator of Tab.~1 of the main paper) and FovVideoVDP compare with the Ultra Flash clips of the same prompts. A good metric agrees with the study, ranking the six conditions as viewers did, so its rank correlation $\rho$ with the user-study JOD approaches the ceiling near 0.8, set by three conditions that lie within 0.03\,JOD of each other. MUSIQ on the $2.5^\circ$ crop and KID on the $20^\circ$ crop agree best. FovVideoVDP is given only for the two conditions that keep Ultra Flash's content.}
  \label{tab:study-metrics}
\end{table*}

\paragraph{Temporal consistency.}
SC, BC and MS~\cite{huang2024vbench}, computed along the moving gaze, include every region shift. Before the viewing filter, \textit{Balanced} lies 0.89 (SC) and 0.91 (BC) points below Ultra Flash. No configuration flickers more than Ultra Flash in VBench temporal flickering, and region switches add a small change at the region's edge and little at the gaze (\cref{sec:supp-temporal}).

\paragraph{Prompt alignment.}
Gaze allocation should keep the prompt's content in the periphery. We therefore measure the text--video similarity of the 158 acuity-filtered clips of Tab.~1 of the main paper with ViCLIP (VBench overall consistency) and CLIPScore (CLIP ViT-L/14), both $\times100$, paired against Ultra Flash over prompts. The ideal is no change from Ultra Flash, and both configurations come close. \textit{Balanced} stays within 0.07 on both scores, and \textit{Fast} stays within 0.15 on ViCLIP but loses 0.15 in CLIPScore (95\% CI [$-0.27$, $-0.02$]), the only difference of ours that excludes zero. Uniform, at the same throughput as \textit{Balanced}, loses more, 0.49 in ViCLIP and 0.23 in CLIPScore, with intervals that exclude zero.

\section{User study}
\label{sec:supp-study-all}
The user study tests whether viewers see the details that get removed by \method{} in the periphery.
\input{figures/tex/study_results}

\subsection{Protocol}
\label{sec:supp-study-protocol}
Eleven participants (7 male and 4 female, aged 21 to 32, mean 24.4), naive to the purpose of the study and with normal or corrected-to-normal vision, gave informed consent under a protocol approved by the university's institutional review board (IRB).
The stimuli use 20 of the 158 test prompts whose Ultra Flash outputs are sharp in the periphery, since a defocused background would hide peripheral differences, meaning that the ratio of the Laplacian variance within $5^\circ$ of the screen center to that beyond $10^\circ$ is below 1.6. Ten come from the 36-prompt subset and ten are the highest-ranked remaining ones (indices 15, 20, 23, 28, 38, 41, 51, 60, 72, 83, 84, 101, 105, 124, 125, 130, 140, 150, 152 and 157, counted from 0).
Each prompt is rendered in the six conditions (Ultra Flash, filtered Ultra Flash, \textit{Balanced}, \textit{Fast}, Latent Upsampling and Uniform, 120 clips), regenerated with every gaze-driven stage reading the screen center as the gaze, as 165 frames at $1664\times960$ from the tiny decoder (TAEHV), encoded with H.264 at CRF 17 and played at 16\,fps (10.3\,s).
Participants rested their head on a chin rest 32.8\,cm from a 2.5K display whose width the clips fill, giving every participant the $47.9^\circ\times28.1^\circ$ field our metrics assume (Sec.~4.1 of the main paper), and were asked to fixate a red cross ($0.6^\circ$) at the screen center, shown before and during both clips.
Each trial is a two-interval forced choice (cross on gray for 1\,s, first clip, cross for another 1\,s, second clip) in randomized clip order. Without practice trials, participants chose the clip with higher overall visual quality, described as sharpness, artifacts, flicker and naturalness, guessing when unsure.
In each of four blocks, ASAP~\cite{mikhailiuk2020active} selected one condition pair for each of the 20 prompts, in randomized prompt order, giving 80 trials per participant in sessions of 35 to 40 minutes with breaks between blocks.

\paragraph{Central fixation and moving gaze.}
Fixing the gaze excludes the hardest cases for a foveated stream, namely a saccade that lands outside the refined region, the later refinement of the new location and artifacts as the region moves, so the study measures only whether the allocation is visible when gaze prediction is correct.
A gaze-contingent study would need live generation at about 20\,fps playback rate or higher, which \method{} misses on one GPU (at most 8.3\,FPS on an L40S and 12.4\,FPS on an H100 at 1K, Tab.~3 of the main paper). Live generation would also break the paired comparison, since RA and KV change the content with the gaze.
We evaluate moving gaze with simulated viewers (\cref{sec:supp-simgaze,tab:gaze-oracle}), whose whole-frame consistency scores would barely register a brief artifact at a region boundary. Whether viewers notice a refinement that arrives after a missed saccade remains to be tested with live eye tracking.

\subsection{Scaling and statistical tests}
\label{sec:supp-study-stats}
We scale the choices, pooled over prompts, with a Python port of pwcmp~\cite{perezortiz2017pwcmp} (Thurstone Case V with a Gaussian prior), fixing Ultra Flash at 0\,JOD. A difference of 1\,JOD means that 75\% of observers prefer the better condition.
Two of the eleven participants were excluded as outliers by pwcmp's leave-one-out criterion ($\text{dist}_L > 1.5$), fixed before collection ended, leaving 9 participants and 720 trials.
The 95\% intervals are the 2.5th and 97.5th percentiles over 2000 bootstrap resamples of observers, with the scale refitted each time. Tests use pwcmp's normal approximation with the variance of each difference from the bootstrap covariance, and two-sided $p$-values are Holm-corrected over all 15 condition pairs.
With the 9 included participants, only Ultra Flash, filtered Ultra Flash and \textit{Balanced} are mutually indistinguishable (uncorrected $p > 0.8$), and every other pair differs at uncorrected $p < 10^{-3}$ and Holm-corrected $p < 0.003$. We do not rank \textit{Fast} against Latent Upsampling, the only difference that depends on the outlier exclusion.
A Kolmogorov--Smirnov test on the bootstrap distributions is consistent with normality ($p > 0.05$) for \textit{Balanced}, \textit{Fast} and filtered Ultra Flash but rejects it for Latent Upsampling and Uniform ($p < 0.002$), whose near-unanimous losses skew their bootstrap distributions. The tests behind our claims involve only conditions that pass, and these two lie more than 1.4\,JOD below both references, with intervals that exclude zero by more than 1\,JOD, so their small $p$-values should be read as very small rather than exact.
\section{Gaze and the commit horizon}

Each foveated stage fixes its region before its frames are shown (Sec.~3.4 of the main paper), so the gaze can leave the region while they play; this section measures how often that happens and what it costs.
Regions and gaze crops map the $1664\times960$ frame linearly onto a $48^\circ\times28^\circ$ field, 34.7 pixels per degree horizontally and 34.3 vertically, while the viewing filter uses pyfvvdp's flat-display model, with 32.9 pixels per degree at the center.

\subsection{Simulated viewers}
\label{sec:supp-simgaze}
Every moving-gaze result uses simulated viewers.
The video gaze model of~\cite{kang2026infinitegaze}, trained on DIEM eye tracking~\cite{mital2011diem}, generates 16 viewers per prompt from UNISAL~\cite{droste2020unisal} saliency features of the Ultra Flash clip, starting at the screen center; viewer~0 drives generation unless noted.
They watched an A6000 render of each prompt with the same seed, not the slightly different L40S render we score, and every method on a prompt is scored along the same trajectory.
Driving each of 36 prompts with one of viewers 1--15 instead changes coverage by $-1.0$ points (paired 95\% CI [$-4.7$, $2.8$]); their DINOv2 similarity to Ultra Flash at the gaze, 0.828 against 0.837 for viewer~0, comes from runs on different GPU types and is not paired.

\subsection{Execution schedule}
\label{sec:supp-schedule}
The commit horizon sets how far ahead each stage must predict the gaze.
RA and the KV memory read the gaze 0.75\,s before an LR chunk's first frame, and the chunk plays for 0.75\,s, a 1.5\,s horizon, which RA covers by widening its region to every gaze sample of the preceding 0.75\,s, with a $10^\circ$ radius. Each SR call reads the gaze 0.25\,s before its first frame and plays for 0.5\,s, a 0.75\,s horizon.
Our experiments run slower than real time, but every stage reads only gaze samples timestamped before its deadline, except the group-wise decoder window of \textit{Fast} (\cref{sec:supp-causal-fast}).
Live, a \textit{Balanced} SR group of 24 frames takes 1157\,ms on one L40S (48.2\,ms per frame for its SR DiT, decoder and other SR work, \cref{tab:supp-profile}), about 386\,ms per 0.5\,s call, longer than the 0.25\,s in which a call must refine its frames and decode the first, so a single L40S cannot meet this schedule live.
Whole-clip results change little with the lead (\cref{tab:supp-sr-sweeps}), but a longer lead costs most in the second after a saccade (a gaze displacement above $5^\circ$ within 0.25\,s): with a 1\,s lead, coverage stays at 71--74\% for that second, and foveal PSNR falls by 0.93\,dB 0.5--1\,s after the saccade, against 0.22\,dB over all frames (36 prompts).

\paragraph{Reading the gaze late, in short calls.}
\label{sec:supp-calls}
We split each 1.5\,s SR call of Ultra Flash into three 0.5\,s calls, each reading the gaze shortly before its frames. In a replay over all 16 viewers (\cref{tab:supp-staging}), this covers the gaze 0.8 points more often than one 1.5\,s call with the heatmap (paired 95\% CI [0.4, 1.3]) and 1.9 points more with the latest gaze sample, which lacks the heatmap's 0.75\,s of history. It also beats stale SR, which uses the gaze the generator committed with, by 2.5 points ([1.9, 3.1]).
In generated clips (36 prompts), 0.5--1\,s after a saccade stale SR covers the gaze 7.5 points less often for viewer~0 and 7.0 for viewers 1--15, but its foveal-PSNR loss (0.41\,dB) resolves for viewer~0 only; over the whole clip it does not differ measurably for viewer~0 and has 1.1 lower gaze-crop KID at $20^\circ$ for viewers 1--15.
One 1.5\,s call also has lower gaze-crop KID at $20^\circ$ (by 2.4 for viewer~0 and 2.0 for viewers 1--15) and no measurable difference at $5^\circ$, but it must refine three times as many frames within the same lead.

\begin{table}[t]
  \centering
  \tablesize
  \setlength{\tabcolsep}{4pt}
  \begin{tabular}{@{}l cc@{}}
    \toprule
    & \multicolumn{2}{c}{Gaze inside SR region (\% of frames)} \\
    \cmidrule(l){2-3}
    SR schedule & Heatmap region & Latest-gaze region \\
    \midrule
    0.5\,s calls (ours) & \textbf{83.7} & \textbf{79.4} \\
    0.75\,s calls & 83.5 & 78.1 \\
    One 1.5\,s call & 82.9 & 77.4 \\
    Stale SR & 81.3 & -- \\
    Stale SR, one 1.5\,s call & 80.3 & -- \\
    \bottomrule
  \end{tabular}
  \caption{\textbf{Gaze coverage by SR schedule} (replay of all 16 simulated viewers on the 158 prompts, frames from 1.3\,s on, which every schedule foveates). The SR region comes from the recent-gaze heatmap or sits at the latest gaze sample. Longer calls keep one region for more frames, and stale SR uses the gaze the generator committed with, 0.75\,s before its chunk, instead of a later read. Higher is better: reading the gaze late, in short calls, covers it most often. Bold marks the best value in each column; --: not evaluated.}
  \label{tab:supp-staging}
\end{table}

\subsection{Decoder window scheduling}
\label{sec:supp-causal-fast}
Foveated decoding in \textit{Fast} decodes at full resolution only inside a window around the SR regions. The window sets only the decoder's resolution, not which blocks SR refines, so SR and every coverage and miss statistic stay causal. The window can be scheduled in two ways. \textit{Fast (group-wise)}, used throughout the paper, decodes each 1.5\,s SR group with one window enclosing the regions of its three 0.5\,s calls, so its earliest frames use region decisions made up to 0.75\,s after their own. \textit{Fast (per-call, causal)} decodes each call's frames with a window around that call's region only, so no frame waits for a later decision. Under central gaze all three calls choose the same region and the two schedules coincide, so the user study is unaffected, as is \textit{Balanced}, which decodes the whole frame.
On 158 prompts (viewer~0, one L40S), the per-call schedule runs as fast (8.36 against 8.26\,FPS over 36 timing prompts) and does not flicker more (VBench temporal flickering 98.25 against 98.22, where higher is steadier). Its smaller window (35.6 against 45.3\% of the frame) contains the gaze in 92.4 instead of 97.4\% of frames, which costs 0.99 points of gaze-crop MUSIQ (95\% CI [$-1.18$, $-0.80$]) and 0.38 points of the acuity-filtered average ([$-0.42$, $-0.34$]). The speed-up of foveated decoding thus does not depend on later region decisions, while about one point of the group-wise schedule's gaze-crop MUSIQ does. Widening the per-call window by dilation, by the two previous calls' regions or by a $5^\circ$ disc at the gaze does not recover this point (36 prompts).

\subsection{Recent-gaze heatmap}
\label{sec:supp-heatmap}
The heatmap decides where SR refines, so we measure how much of the gaze it catches.
In a replay of viewer~0's gaze on the 158 prompts at our 22\% budget, it covers 84.9\% of gaze samples, against 80.3\% for a region at the latest gaze sample, 66.6\% for a fixed central region and 100\% for an oracle that knows the gaze during the call's display, so the remaining misses come from predicting the gaze, not from the budget.
The latest sample covers more at a 10\% budget (65.0\% against 62.6\%), and a history longer than our 0.75\,s covers slightly more (86.3\% at 3\,s).

\paragraph{KV retention over time.}
At the same 25\% budget (36 prompts), gaze-driven KV gains 1.29\,dB of gaze-crop PSNR over the sliding window against a full-cache reference with the same re-indexing, 2.27\,dB in the first 5\,s, with an interval that includes zero after 5\,s; the sliding window scores 0.66 points higher in gaze-crop MUSIQ, mostly after 5\,s.

\subsection{Region placement, size and latency}
\label{sec:supp-placement-sweeps}
The SR region's placement, size and lead set the trade-off between speed and quality at the gaze.
These sweeps vary foveated SR alone, which keeps the LR video fixed, so foveal PSNR and the gaze-crop MUSIQ change are paired against Ultra Flash from the same run.

\paragraph{Placement.}
At the same 22\% budget, the heatmap covers the most gaze samples and gives the highest foveal PSNR of the causal placements, below a $5^\circ$ disc at the true gaze (\cref{tab:gaze-oracle}, top).
Super-resolving a hard-edged $5^\circ$ crop at the gaze without the surrounding frame lowers gaze-crop DINO similarity from 0.967 to 0.961 (paired 95\% CI of the change [$-0.008$, $-0.005$]), so foveated SR keeps the whole frame as context.

\begin{figure}[t]
  \centering
  \includegraphics[width=\linewidth]{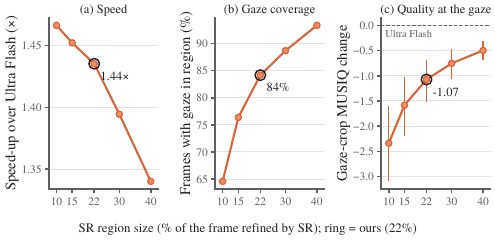}
  \caption{\textbf{How large should the SR region be?} Foveated SR alone with the tiny decoder, for SR regions of 10 to 40\% of the frame. A larger region costs speed (a; six timing prompts on one L40S) but contains the gaze in more frames (b) and loses less detail at the gaze (c; gaze-crop MUSIQ minus Ultra Flash's, 95\% prompt-bootstrap intervals); (b) and (c) use 36 prompts. Our 22\% (ring) is the knee: shrinking the region to 10\% gains only 0.03$\times$ of speed-up but drops coverage from 84 to 65\%, while growing it to 40\% raises coverage to 93\% at a cost of about 0.1$\times$.}
  \label{fig:region-share}
\end{figure}

\paragraph{Region size.}
A larger SR region covers the gaze more often and keeps more quality at the gaze, but runs slower (\cref{fig:region-share,tab:supp-sr-sweeps}). Our 22\% sits at the knee: 84\% coverage at 1.44\x{}, against 65\% at 1.47\x{} for 10\% and 93\% at 1.34\x{} for 40\%.

\paragraph{Lead and history.}
The lead $\Delta$ is the time between choosing the SR region and displaying its first frame (Sec.~3.4 of the main paper). A longer lead lowers coverage and foveal PSNR at the same speed, and a heatmap history longer than our 0.75\,s raises the MUSIQ change slightly (\cref{tab:supp-sr-sweeps}).

\begin{table}[t]
  \centering
  \tablesize
  \setlength{\tabcolsep}{2.2pt}
  \begin{tabular}{@{}l ccccc@{}}
    \toprule
    & Gaze in & Speed- & Foveal & MUSIQ & Gaze-crop \\
    Setting & region (\%)$\uparrow$ & up$\uparrow$ & PSNR (dB)$\uparrow$ & change$\uparrow$ & KID$\downarrow$ \\
    \midrule
    \multicolumn{6}{@{}l}{\emph{Region size (share of SR blocks)}} \\
    10\% & 65 & 1.47\x & 30.55 & $-2.34$ & 18.34 \\
    22\% (ours) & 84 & 1.44\x & 33.13 & -- & 6.11 \\
    40\% & 93 & 1.34\x & 34.47 & $-0.50$ & 4.13 \\
    \midrule
    \multicolumn{6}{@{}l}{\emph{Lead $\Delta$ from choosing the region to its first frame}} \\
    0\,ms & 89 & 1.44\x & 33.40 & -- & -- \\
    250\,ms (ours) & 84 & 1.44\x & 33.19 & -- & -- \\
    1\,s & 81 & 1.44\x & 33.01 & -- & -- \\
    \bottomrule
  \end{tabular}

  \vspace{4pt}
  \setlength{\tabcolsep}{3pt}
  \begin{tabular}{@{}l cccc@{}}
    \toprule
    Heatmap history $T$ & 0.25\,s & 0.75\,s (ours) & 1.5\,s & 3\,s \\
    MUSIQ change$\uparrow$ & $-1.32$ & $-1.19$ & $-1.13$ & $-1.13$ \\
    \midrule
    Lead $\Delta$, 158 prompts & 0.25\,s (ours) & 0.375\,s & 0.875\,s & \\
    Gaze in region (\%)$\uparrow$ & 84.0 & 83.1 & 81.6 & \\
    \bottomrule
  \end{tabular}
  \caption{\textbf{Region size, lead and history for foveated SR alone} (tiny decoder, 36 prompts unless noted). A larger region raises quality at the gaze but lowers the speed-up; a shorter lead raises coverage at the same speed, but live SR needs time to run (\cref{sec:supp-schedule}). Gaze in region: share of gaze samples inside the SR region. Foveal PSNR and the gaze-crop MUSIQ change are against Ultra Flash from the same run; Ultra Flash's own gaze-crop KID is 4.09 in the size sweep and 5.35 in the lead sweep. Speed-up is from timing-only runs over 6 prompts on one L40S. Over all five leads from 0 to 1\,s, the MUSIQ change lies between $-1.25$ and $-1.10$ and gaze-crop KID between 6.89 and 7.51. The history sweep uses a 250\,ms lead; the 158-prompt lead replay covers all 16 simulated viewers per prompt.}
  \label{tab:supp-sr-sweeps}
\end{table}

\subsection{Gaze prediction against the true gaze}
\label{sec:supp-gaze}
An oracle that knows the future gaze bounds what better gaze prediction could gain.
True gaze here is the simulated viewer's evaluation track, including samples unavailable at the causal deadlines, not measured human gaze.
\Cref{tab:gaze-oracle} varies only the SR placement (top) or runs every stage of \textit{Fast} on causal or true gaze (bottom). In the oracle, the SR region is a $5^\circ$ disc with a $1^\circ$ margin that follows the gaze of each 4-frame block, RA uses the gaze recorded while the chunk is on screen, and KV retention uses the larger of the current gaze weight and the weight 0.75\,s later.
The oracle gains 0.83 points of gaze-crop MUSIQ (paired 95\% CI [0.22, 1.47]) at similar throughput, and its paired KID difference includes zero.

\begin{table}[t]
  \centering
  \tablesize
  \setlength{\tabcolsep}{2.2pt}
  \begin{tabular}{@{}l ccccc@{}}
    \toprule
    & Gaze in & Speed- & Foveal & MUSIQ & Gaze-crop \\
    Gaze source & region (\%)$\uparrow$ & up$\uparrow$ & PSNR (dB)$\uparrow$ & change$\uparrow$ & KID$\downarrow$ \\
    \midrule
    \multicolumn{6}{@{}l}{\emph{Foveated SR alone, region from}} \\
    Heatmap (ours) & \textbf{84} & 1.44\x & \textbf{33.01} & \textbf{\boldmath$-1.03$} & 3.88 \\
    Latest sample & 80 & 1.44\x & 32.77 & $-1.17$ & \textbf{3.05} \\
    Latent saliency & 56 & 1.44\x & 31.22 & $-1.86$ & 11.39 \\
    Screen center & 62 & 1.44\x & 31.59 & $-1.74$ & 9.27 \\
    True-gaze disc & 100 & 1.44\x & 34.22 & $-0.57$ & 3.77 \\
    \midrule
    \multicolumn{6}{@{}l}{\emph{\textit{Fast}, all stages on}} \\
    Causal gaze & 84 & 1.82\x & -- & $-2.20$ & 3.82 \\
    True gaze & 100 & 1.83\x & -- & $-1.37$ & 2.44 \\
    \bottomrule
  \end{tabular}
  \caption{\textbf{Gaze prediction against the true gaze} (36 prompts, released tiny decoder). Higher coverage, PSNR and MUSIQ change, and lower KID, are better. Top: foveated SR alone in 0.5\,s calls. Among causal sources the heatmap covers the gaze most often and gives the highest PSNR and MUSIQ change, though the latest gaze sample has lower KID; only a $5^\circ$ disc at the true gaze, which needs future gaze, scores higher. PSNR and the MUSIQ change are against Ultra Flash from the same run, and gaze-crop KID ($\times10^3$) against four independent Ultra Flash seeds (Ultra Flash: 0.32). Bottom: \textit{Fast} with every stage on causal or true gaze, viewer~0, generated on an RTX A6000. The MUSIQ change is against Ultra Flash's 61.58 from the same job, KID is the unbiased estimate on $10^\circ$ gaze crops, and PSNR is omitted because RA and KV change the content. Speed-ups are from timing-only L40S runs, over six prompts (top) or the same 36 prompts (bottom; Ultra Flash 4.43\,FPS). Bold marks the best value in each column within each group, excluding the true-gaze rows.}
  \label{tab:gaze-oracle}\label{tab:supp-gaze}
\end{table}

\subsection{When the gaze leaves the region}
\label{sec:supp-misses}
A saccade can carry the gaze outside the SR region before the next SR call moves the region. These misses are where foveation could become visible, so we measure how often they happen, how long they last and what they cost, over the 25{,}280 frames after the first SR group of 158 prompts, with viewer~0's gaze. An \emph{excursion} is a run of frames with the gaze outside the region, and a \emph{gaze shift} is a jump of more than $1.5^\circ$ between consecutive gaze samples at 16\,fps. \Cref{fig:supp-timestrip} shows two such misses.

\paragraph{Occurrences.}
The gaze lies outside the SR region in 15.6\% of frames, and outside the wider region of region-adaptive denoising in 5.5\%. Of all gaze shifts, 74\% land inside the current SR region.

\paragraph{Duration.}
Misses are short. The median excursion lasts 0.25\,s, 89\% end within 0.5\,s and 97\% within 0.75\,s, and the longest lasts 1.8\,s. When a shift lands outside the region, in half of the cases the viewer moves on before the next SR call; in the other half a later call refines the new location, after a median of 0.25\,s (90th percentile 0.44\,s). Coverage therefore recovers quickly after a shift: the gaze lies inside the SR region in 74\% of frames in the first 0.25\,s after the shift, 82\% from 0.25 to 0.5\,s, 93\% from 0.5 to 1\,s and 98\% after 1\,s. Once the viewer dwells long enough to inspect a location, the refined region is almost always there.

\paragraph{Cost.}
Outside the region, the viewer sees the upsampled latent, with about the detail of Latent Upsampling (Tab.~1 of the main paper), until a later call refines the new location. With foveated SR alone, gaze-crop PSNR against Ultra Flash in 0.5\,s calls is 34.84\,dB while the gaze is inside the region and 28.53\,dB while it is outside, a drop of about 6.3\,dB. In the full pipeline, a frame in which SR misses the gaze but region-adaptive denoising still covers it loses 6.6\,dB of PSNR and 0.070 of LPIPS (on the $5^\circ$ crop, against the same generator with full-frame SR) and 0.76 points of gaze-crop MUSIQ (on the $2.5^\circ$ crop, as the change from Ultra Flash in 0.5\,s calls), compared with frames in which both regions hold the gaze. The reverse case, a miss of region-adaptive denoising that SR still covers, costs 0.20 MUSIQ points, which is not significant: region-adaptive denoising reads the gaze earlier than SR but covers a wider region, so the visible cost of a miss comes from SR, the stage with the shorter horizon.

\subsection{Temporal stability at region shifts}
\label{sec:supp-temporal}
A moving region could add flicker that per-frame metrics miss.
On frames where a new call moves the SR region, we measure the rise in the frame-to-frame change $|\mathbf e_t-\mathbf e_{t-1}|$ of the error between foveated SR and the same generator with full-frame SR, which share the LR video (0--255 scale), net of the rise of about 0.5 beyond $20^\circ$, which no switch can reach.
Switches add a small, measurable change at the region's edge ($+0.32$ for foveated SR, $+0.33$ on \textit{Balanced}'s generator), about 5\% of the reference's own frame-to-frame change there (6.2), and little at the gaze ($+0.05$, whose interval includes zero, and a resolved $+0.09$ on \textit{Balanced}'s generator).
In VBench temporal flickering on the 158 clips of Tab.~1 of the main paper (higher means less flicker), no configuration flickers more than Ultra Flash, which scores 98.24 after the acuity filter: \textit{Balanced} scores $+0.14$ higher and \textit{Fast} $+0.22$ ($+0.17$ and $+0.34$ unfiltered).
These gains grow with the loss of fine detail outside the region, so they show no flicker from the moving region rather than a gain in temporal quality, as for the causal window of \textit{Fast} and foveated FlashVSR (\cref{sec:supp-causal-fast,sec:supp-flashvsr}).

\subsection{Closed-loop gaze}
\label{sec:supp-closedloop}
The simulated viewer watches the Ultra Flash clip, so its gaze G$_0$, which drives every variant, cannot react to our output.
To close the loop, iteration $k$ generates \method{} with gaze G$_{k-1}$, and the viewer, with the same seed and noise, watches that output to give G$_k$ (36 prompts, viewers 0--3, tiny decoder, L40S).
The mean distance between gaze tracks is $4.53^\circ$ in the first iteration and $2.76^\circ$ in the second, about as much as switching between full and foveated decoding in the same run moves the gaze ($2.80^\circ$), and the total shift, $4.66^\circ$ (95\% CI [4.06, 5.31]), stays below the $6.08^\circ$ ([5.31, 6.91]) between viewers of the same clip.
Scored against the gaze the pipeline read rather than the viewer's response, coverage is 5.2 points higher in the first iteration and 2.4 in the second.
In the second iteration, the gaze is inside the SR region on 82.4\% of foveated frames against 80.4\% of full-decode frames, which may hint that sharp regions attract gaze, but one iteration is not enough evidence for a claim.

\section{KV memory}
Gaze KV keeps 25\% of the generator's cache (Sec.~3.3 of the main paper). This section gives its saliency head and implementation, tests the terms of its retention score, and shows how fidelity and memory change with the budget.

\subsection{Latent saliency head}
\label{sec:supp-saliency}
The saliency term $S_c$ of the retention score (Eq.~(2) of the main paper) comes from a small head on the LR latents, the only trained component of \method{}; only the KV score uses it ($w_{\mathrm s}=0.5$).

\paragraph{Architecture.}
The head has 0.48M parameters and runs in 0.7\,ms per latent frame. It takes the LR latents at frames $t$ and $t-1$ (zeros at $t=0$), 32 channels on the $60\times104$ latent grid, and applies a $3\times3$ convolution to 64 channels, three residual blocks of two GroupNorm--SiLU--$3\times3$ convolutions, a $2\times2$ convolution with stride 2 to the $30\times52$ token grid, three more residual blocks, and a GroupNorm--SiLU--$1\times1$ convolution to one logit per token, followed by a spatial softmax.

\paragraph{Training.}
The head learns saliency distilled from UNISAL~\cite{droste2020unisal}, with no human gaze. Its 408 clips are Ultra Flash release rollouts of the 158 prompts (LR latents only, 21 latent frames each) and Matrix-Game~2.0~\cite{he2025matrixgame2} rollouts, split at random into 368 for training and 40 for validation. The target of a latent frame is UNISAL's saliency (DHF1K domain) of the output frames it decodes to, averaged over the four (one for the first latent), area-pooled to the token grid and normalized to sum to one. Because the Ultra Flash clips include the test prompts, the head has seen their release LR latents and UNISAL's saliency for them, but no gaze, no quality signal and none of our method's outputs.

\paragraph{Accuracy.}
On the 40 validation clips, the predicted maps correlate with UNISAL's at 0.75 (0.71 on Ultra Flash clips, 0.77 on Matrix-Game clips), against 0.57 for a center-bias Gaussian fitted to the training targets; this is agreement with UNISAL, not with human fixations. As a gaze predictor the head is weaker: used as the SR region on our 36 prompts, it covers 56\% of the simulated viewer's gaze samples, against 84\% for the recent-gaze heatmap (\cref{tab:gaze-oracle}). SR therefore does not use it, and in the KV score it adds only a weak content prior.

\paragraph{Score ablation.}
At the 25\% budget on 158 prompts, removing the saliency or the past-gaze term ($w_{\mathrm s}=0$ or $w_{\mathrm p}=0$) does not measurably change gaze-crop fidelity (PSNR $-0.03$ and $-0.01$\,dB, LPIPS $+0.001$ for both); only the MUSIQ change without the past-gaze term, $+0.15$ (95\% interval [$0.00$, $+0.30$]), reaches the edge of its interval, in favor of removing the term.
Even where the terms should matter, in the 0.75\,s after the 209 of 3288 saccades that land more than $5^\circ$ from every earlier gaze sample (saliency) or after the other 3079, which land within $5^\circ$ of earlier gaze (past gaze), no change from removing either term is resolved, so the current-gaze term carries the policy and the saliency head is not needed for the reported fidelity.

\subsection{Implementation}
\label{sec:supp-kv-impl}
\begin{figure*}[t]
  \centering
  \includegraphics[width=\textwidth]{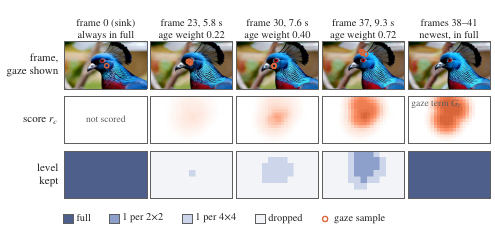}
  \caption{\textbf{What the gaze-driven KV memory keeps} in one rollout (the crowned-pigeon prompt, simulated viewer 0, 25\% budget) after the chunk that ends at latent frame 41: the sink frame, three older cached frames with their age weight $\alpha$, and the four newest latent frames. Rows show each frame with the gaze samples recorded while it was shown, the retention score $r_c$ of Eq.~(2) of the main paper (the current gaze term $G_c$ over the newest frames, which are kept in full), and the level at which each cell is kept. The sink and the four newest frames take 7800 of the 8190 tokens per layer; the policy places the other 390 in the 15 older frames (1.7\% of their tokens), all pooled (184 at 1 per $2\times2$, 206 at 1 per $4\times4$, none in full) and all within $13^\circ$ of the current gaze, here on the bird's head. Because $w_{\mathrm g}$ is twice $w_{\mathrm p}$ and $w_{\mathrm s}$, older frames keep detail mainly where the viewer looks now.}
  \label{fig:kv-anatomy}
\end{figure*}

\Cref{fig:kv-anatomy} shows what the cache keeps in one rollout; this subsection gives what a reimplementation needs beyond Sec.~3.3 of the main paper.

\paragraph{Tokens per layer.}
Each attention layer of Ultra Flash attends to a full cache of 32760 tokens: 21 latent frames of 1560 tokens, 18 cached and the three of the current chunk. The budget $B$ is the number of tokens kept between chunks. It always includes the sink frame and the four newest latent frames in full (7800 tokens), and the retention score places only the rest: 390 of 8190 tokens at 25\%, 2028 of 9828 at 30\%, 5304 of 13104 at 40\% and 8580 of 16380 at 50\%. At 25\%, the 390 movable tokens stand in for up to 1560 or 6240 original positions (one token per $2\times2$ cell or per $4\times4$ block), the equivalent of one or four frames, and the next chunk attends to the 8190 kept tokens plus its own 4680, 39\% of Ultra Flash. All five policies of Tab.~4 of the main paper share everything else and differ only in these movable tokens, which makes the comparison strict and limits how far the policies can differ.

\paragraph{Selection and compaction.}
To find the retention threshold $\lambda$, we evaluate 64 thresholds, spaced logarithmically from $10^{-6}$ to four times the highest cell score, in one batched pass, then 64 more between the last over the budget and the first within it, and use the smallest that fits; this needs two host synchronizations per chunk, where a bisection needs one per step. Cells only move to coarser levels, because a cell pooled at an earlier chunk has already discarded its other entries. One plan serves all 30 layers: it is computed on a side stream during the clean pass, and the compaction into one preallocated buffer overlaps the next chunk. A pooled entry keeps the key and value of its top-left member, with the log-bias of its count, instead of the count-weighted mean.

\paragraph{Attention.}
The log-bias (Eq.~(3) of the main paper) is added inside the fused softmax by a FlashAttention-2 (version 2.8.3) forward kernel patched to add one bias per key: $\log m$ for retained history tokens and zero for the chunk being generated. Pooled tokens come first in the cache, so the kernel skips the bias after that prefix, and one bias vector serves every layer and denoising step while a plan holds. With a zero bias, its output is identical to FlashAttention's. H2O and SnapKV need attention weights, which we recompute outside the fused kernel for 256 strided queries per layer during the clean pass.

\paragraph{Per-chunk policy cost.}
Cache selection and compaction take 18--25\,ms per 0.75\,s chunk for every policy (17.7\,ms for Gaze KV), so they do not explain the throughput differences in Tab.~4 of the main paper.
In a timing rerun on one L40S (six prompts), Gaze KV slows an LR chunk from 898\,ms with the sliding window to 1016\,ms, and still takes 1014\,ms with its policy run synchronously and 1009\,ms without the causal gaze path; of this 118\,ms gap, the log-bias kernel (about 9\,ms) and the 12870 instead of 12480 keys per attention call (about 8\,ms) account for about 17, and we have not located the rest.

\subsection{Token budget}
\label{sec:supp-kv-budget}
The 25\% budget is one point on a trade-off between memory and fidelity, so we test whether the lead of Gaze KV is resolved at 25\% and whether it holds at larger budgets.

\paragraph{Paired intervals for Tab.~4 of the main paper.}
\label{tab:supp-stage-cis}
Over its 158 prompts (prompt bootstrap, 10{,}000 resamples, on the gaze crop; PSNR and LPIPS against the same-job Ultra Flash clip, KID against the pooled release seeds), Gaze KV leads the other 25\% policies in PSNR by $+0.96$ [$0.78$, $1.13$]\,dB over the sliding window, $+0.71$ [$0.55$, $0.87$] over StreamingLLM, $+0.33$ [$0.16$, $0.50$] over SnapKV and $+0.26$ [$0.10$, $0.42$] over H2O. Its LPIPS lead is resolved over the sliding window ($-0.017$ [$-0.024$, $-0.010$]) and StreamingLLM ($-0.015$ [$-0.021$, $-0.008$]) but not over SnapKV ($-0.003$ [$-0.009$, $0.004$]) or H2O ($-0.001$ [$-0.007$, $0.005$]), and no KID difference ($\times10^3$) excludes zero: $-0.01$ [$-1.99$, $1.91$], $-0.10$ [$-2.02$, $1.86$], $+0.12$ [$-1.59$, $1.88$] and $+0.40$ [$-1.21$, $2.09$] in the same order. Against uniform skip, which removes the last two denoising steps everywhere, RA gains $+1.77$ [$1.56$, $1.99$]\,dB of PSNR and $-0.050$ [$-0.058$, $-0.042$] of LPIPS, while its KID difference, $-1.24$ [$-2.87$, $0.30$], includes zero.

To show how the budget trades memory for fidelity, we raise the budget of Gaze KV to 30, 40 and 50\% with all else unchanged (\cref{tab:supp-kv-budget}). From 30 to 50\%, gaze-crop PSNR rises from 22.46 to 25.27\,dB and LPIPS falls from 0.217 to 0.171, while peak memory grows only from 24.5 to 25.7\,GB, so the budget can be set to fit the GPU.

\begin{table}[t]
  \centering
  \tablesize
  \setlength{\tabcolsep}{3pt}
  \resizebox{\columnwidth}{!}{%
  \begin{tabular}{@{}l ccc c@{}}
    \toprule
    & \multicolumn{3}{c}{Gaze crop} & Memory \\
    \cmidrule(lr){2-4}
    Policy and budget & PSNR (dB)$\uparrow$ & LPIPS$\downarrow$ & KID $\times10^3\downarrow$ & (GB)$\downarrow$ \\
    \midrule
    Ultra Flash, 0.5\,s SR calls & 37.38 & 0.012 & 0.65 & 27.9 \\
    \midrule
    Gaze KV, 30\% & 22.46 & 0.217 & 1.54 & 24.5 \\
    Gaze KV, 40\% & 24.01 & 0.190 & 1.27 & 25.1 \\
    Gaze KV, 50\% & 25.27 & 0.171 & 0.96 & 25.7 \\
    \bottomrule
  \end{tabular}}
  \caption{\textbf{Gaze KV at larger token budgets} (158 prompts, viewer~0, one L40S, tiny decoder). PSNR and LPIPS compare the gaze crop with the L40S Ultra Flash clips of Tab.~1 of the main paper, KID compares gaze crops with the pooled release seeds, and memory is the mean peak. A larger budget brings the gaze crop closer to Ultra Flash for about 0.6\,GB per 10 points of budget. The Ultra Flash row is the 0.5\,s-call baseline of Tab.~4 of the main paper on these clips.}
  \label{tab:supp-kv-budget}
\end{table}

\subsection{Temporal position reindexing}
\label{sec:supp-washout}
We reindex cached temporal positions as described in Sec.~3.3 of the main paper. Here we record the settings used in the comparisons and the measured effect on gaze-crop MUSIQ.
The configurations of \method{} and every cache policy in Tab.~4 of the main paper reindex positions; Ultra Flash, Uniform and Latent Upsampling keep the released positions, and the KV comparison over time in \cref{sec:supp-heatmap} uses a full-cache reference with the same reindexing.
Reindexing does not raise the scores. On Ultra Flash's full cache (36-prompt subset, tiny decoder), it leaves gaze-crop MUSIQ unchanged in the first 5\,s ($0.00$) and changes it by $-1.26$ afterwards (95\% interval [$-2.67$, $-0.15$]; $-0.65$ over the whole clip), so comparisons of \method{} with Ultra Flash include this cost.

\section{Region-adaptive denoising}
\label{sec:supp-ra}
RA leaves the final denoising steps undone away from the gaze, and the clean pass caches that less refined periphery, so the schedule decides how far the video drifts from full generation.
Restricting steps on every chunk drifts far from full generation, and a fully denoised chunk every two or three chunks resets the drift (\cref{tab:supp-ra-schedule}). Ours is the fastest schedule closer to full generation than the 25\% KV budget alone.
Keeping the periphery's KV entries from the last full chunk instead of recomputing them is worse (FD 50 against 29).

\begin{table}[t]
  \centering
  \tablesize
  \setlength{\tabcolsep}{2.5pt}
  \resizebox{\columnwidth}{!}{%
  \begin{tabular}{@{}ll cccc@{}}
    \toprule
    Restricted steps & On chunks & ms/chunk$\downarrow$ & Speed-up$\uparrow$ & DINOv2$\uparrow$ & FD$\downarrow$ \\
    \midrule
    None (full generation) & -- & 1049 & -- & -- & -- \\
    Last step & every & 933 & 11\% & 0.871 & 196 \\
    Last step & alternate & 1009 & 4\% & 0.966 & 29 \\
    Last step & 2 of 3 & 983 & 6\% & 0.963 & 33 \\
    Last 2 steps & alternate & 946 & 10\% & 0.950 & 49 \\
    Last 2 steps & 2 of 3 (ours) & 911 & 13\% & 0.941 & 54 \\
    \bottomrule
  \end{tabular}}
  \caption{\textbf{Region-adaptive schedules}: which denoising steps are restricted to the gaze region, and on which chunks, in a six-prompt pilot over 21 latent frames. Time is per LR chunk, and speed-up is against full generation. DINOv2 similarity at the gaze (higher is closer) and the Fr\'echet distance (FD) in DINOv2 features (lower is closer) compare each schedule with full generation. Restricting every chunk drifts far (FD 196); ours is the fastest schedule that stays closer to full generation than the 25\% KV budget alone, which scores about 95 on the same FD.}
  \label{tab:supp-ra-schedule}
\end{table}

Around our setting (36 prompts, RTX A6000, tiny decoder, protocol of Tab.~4 of the main paper), the radius and the number of restricted steps each trade speed for gaze-crop fidelity monotonically, and ours are middle points, not optima. Ours runs 1.12\x{} faster than Ultra Flash in 0.5\,s calls. Relative to it, a $5^\circ$ radius gives 1.15\x{} and changes gaze-crop PSNR by $-1.20$\,dB, and a $15^\circ$ radius gives 1.08\x{} and $+1.93$\,dB; restricting the final 1 or 3 steps instead of 2 gives 1.08\x{} and $+2.45$\,dB or 1.17\x{} and $-2.04$\,dB. LPIPS moves with PSNR, and all these PSNR and LPIPS changes exclude zero, but gaze-crop MUSIQ does not separate the settings.
Gaze quality does not drift over the clip. On 158 prompts, \textit{Fast}'s gaze-crop MUSIQ difference to Ultra Flash changes by $-0.17$ [$-1.07$, $+0.71$] from the first to the last 2.5\,s window.

\section{Foveated SR and decoding}
\label{sec:supp-sr-decode}
These details complete Secs.~3.6 and~3.7 of the main paper for a reimplementation. The SR region covers 22\% of the frame's 104 blocks of $8\times8$ tokens (23 blocks) and is blended into the upsampled frame over a $1^\circ$ band (3 tokens, $1.4^\circ$). The SR cache keeps the released $9\times9$-block attention window with up to 3 time blocks and its top-$k$. The decoder window adds a margin of 4 latent pixels, is aligned to 4 latent pixels, and is feathered over 32 pixels with a smoothstep.
Blocks are ranked by their highest token score. Only refined blocks compute queries, projections, cross-attention and feed-forward layers. An unselected block computes nothing and keeps the content of the last call that refined it. The first SR group runs in full.
Super-resolving a hard-edged gaze crop without the surrounding frame lowers fidelity at the gaze (\cref{sec:supp-placement-sweeps}).

\section{Additional qualitative comparisons}
\label{sec:supp-qualitative}
\Cref{fig:supp-qualitative} examines whether concentrating computation near the gaze preserves useful detail across different types of content. Seven additional scenes cover animals, people, animation, and landscapes, with Ultra Flash and our two configurations placed beside the external baselines. Shared gaze crops expose local texture, while equally sized peripheral crops show where computation is reduced. In these examples, both configurations retain detailed gaze regions, whereas \textit{Fast} more visibly softens the periphery. The comparison illustrates a spatial quality trade-off rather than uniform degradation across the frame; differences in generated content prevent pixelwise comparisons between systems.
\input{figures/tex/supp_qualitative}

\section{Qualitative results over time}
\label{sec:supp-timestrip}
\Cref{fig:supp-timestrip} shows two scenes in which the gaze briefly leaves the sharp SR region of \textit{Balanced} and \textit{Fast} around a saccade. While the gaze is outside, the viewer sees the latent upsampler's output there. It is softer than SR but keeps the same content, so quality at the gaze drops a little rather than breaking down, until the gaze returns to the region or the next SR call moves the region to the gaze.

\Cref{sec:supp-misses} measures this over 158 prompts. When the gaze is outside the SR region but inside the wider region of region-adaptive denoising, gaze-crop MUSIQ is 0.76 points lower (on a 0--100 scale) than when the gaze is inside both. PSNR, which compares every pixel with the full-frame SR output, drops more, by 6.6\,dB. The median miss lasts 0.25\,s. The figure shows still frames, so it shows what a miss looks like, not whether a viewer would notice it in motion.
\input{figures/tex/supp_timestrip}

\input{sec/X_suppl_experiments}

%% file: figures/tex/speedVsQuality.tex
\begin{figure}[t]
  \centering
  \includegraphics[width=\linewidth]{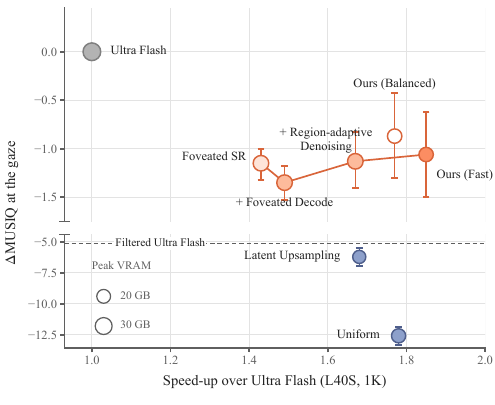}
  \caption{\textbf{Speed against quality at the gaze} for the rows of Tab.~1 of the main paper (TAEHV decoder): speed-up over Ultra Flash (one L40S, 36 prompts) against the change in MUSIQ on the $10^\circ$ gaze crop (158 prompts; 0 is Ultra Flash, higher is better), with 95\% bootstrap intervals over prompts. Marker area shows peak memory, and the connected points add our components one at a time.
  \textit{Balanced} and \textit{Fast} lose about one MUSIQ point at 1.77 and 1.85\x{} the speed, while Uniform at the same speed loses 12.59 and Latent Upsampling 6.21. For scale, applying the viewing filter to Ultra Flash itself lowers gaze-crop MUSIQ by 5.14 (dashed line). }
  \label{fig:speed-quality}
\end{figure}

%% file: figures/tex/study_results.tex
\begin{figure}[t]
  \centering
  \includegraphics[width=\linewidth]{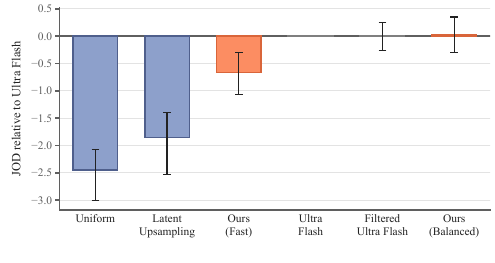}
  \caption{\textbf{User study results} (9 participants, 720 trials; 11 tested, 2 excluded by pwcmp's outlier criterion, see supplementary). Perceived quality in JOD relative to Ultra Flash for the conditions of Tab.~1 of the main paper and Ultra Flash with the acuity filter of Sec.~4.1 of the main paper, with 95\% bootstrap intervals over observers. Two-interval forced choice with ASAP pair selection, 20 prompts, central fixation, $47.9^\circ\times28.1^\circ$ field.
  Relative to Ultra Flash, \textit{Balanced} scores $+0.03$ JOD [$-0.31$, $0.35$], \textit{Fast} scores $-0.67$ [$-1.06$, $-0.30$], and filtered Ultra Flash scores $+0.02$ [$-0.27$, $0.25$].}
  \label{fig:user-study}
\end{figure}

%% file: figures/tex/supp_qualitative.tex
\begin{figure*}[p]
  \centering
  \includegraphics[width=\textwidth,height=0.88\textheight,keepaspectratio]{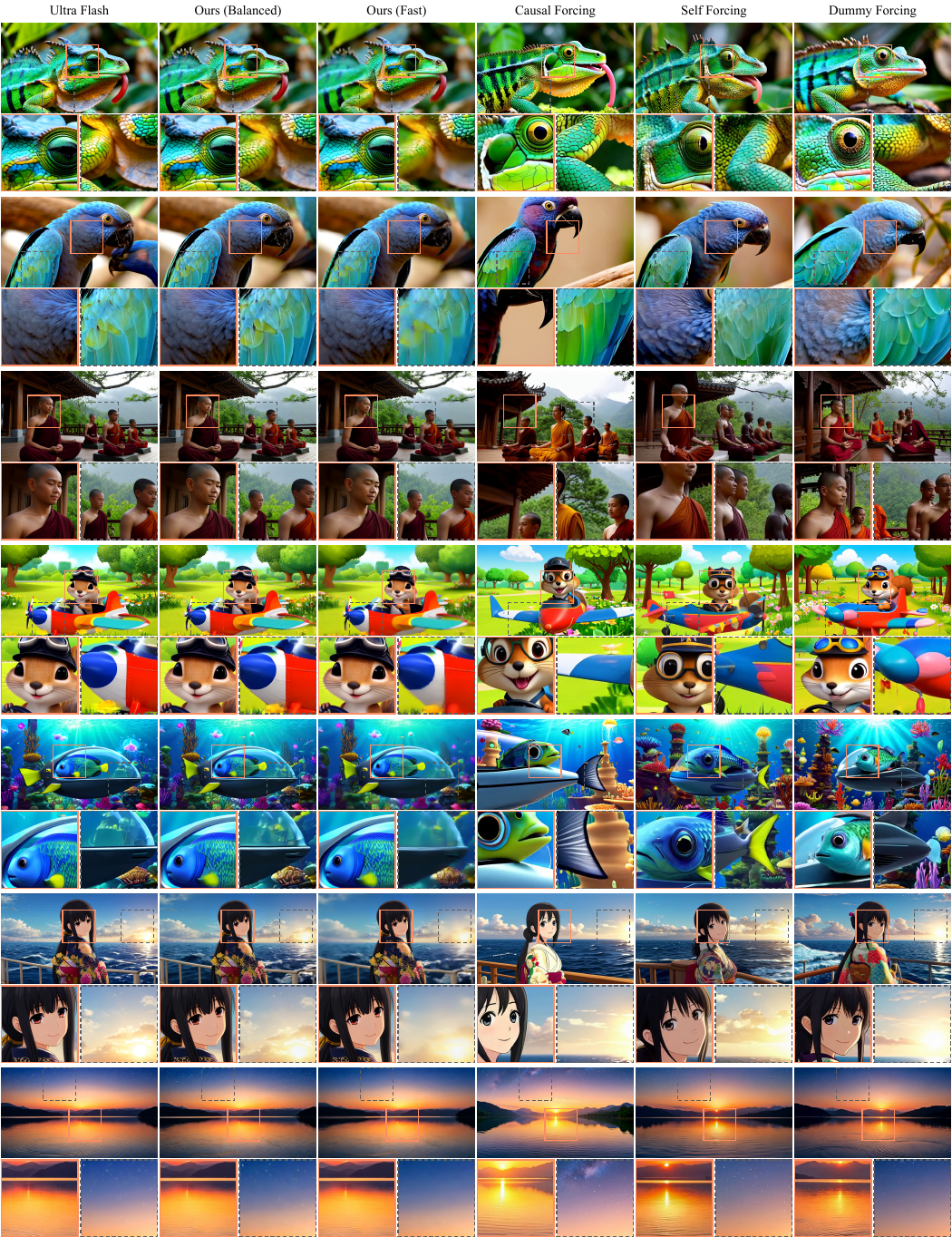}
  \caption{\textbf{More qualitative comparisons.} Seven scenes at the same playback time, without the acuity filter. Below each full frame are its gaze crop (solid \textcolor[HTML]{FC8D62}{orange}) and a peripheral crop (dashed \textcolor[HTML]{5F5F5F}{gray}). At the gaze, \textit{Balanced} and \textit{Fast} keep detail close to Ultra Flash and generally finer than the other generators; in the periphery, where acuity is low, they are softer, \textit{Fast} most visibly. Generated content can differ between methods.}
  \label{fig:supp-qualitative}
\end{figure*}

%% file: figures/tex/supp_timestrip.tex
\begin{figure*}[p]
  \centering
  \includegraphics[width=\textwidth,height=0.84\textheight,keepaspectratio]{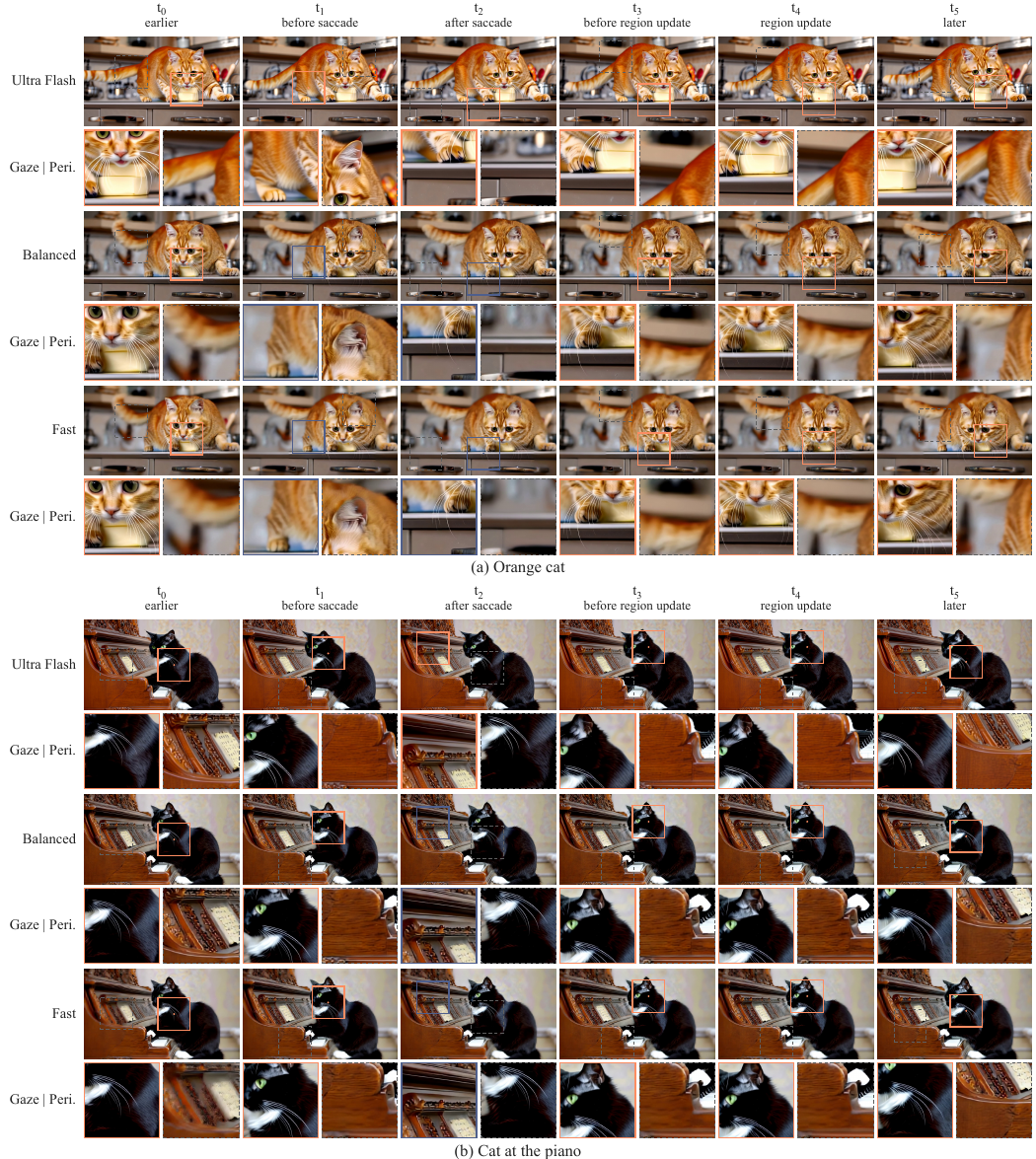}
  \caption{\textbf{Quality at the gaze when it leaves the sharp region} (unfiltered frames). Each scene is centered on its largest one-frame gaze jump, a saccade of $7.3^\circ$ and $16.8^\circ$, chosen without looking at quality. The rows show Ultra Flash, \textit{Balanced} and \textit{Fast} at the same moments, with the same gaze. Under each frame are a crop at the gaze and a crop in the periphery (dashed \textcolor[HTML]{5F5F5F}{gray}). The gaze crop is outlined in \textcolor[HTML]{FC8D62}{orange} when the gaze point is inside the sharp SR region and in \textcolor[HTML]{4F5F8C}{blue} when it is outside (only the gaze point is checked, not the whole crop). In the blue crops, \textit{Balanced} and \textit{Fast} show the latent upsampler's output, as Latent Upsampling does: softer than SR, but with the same content. The columns are: \emph{earlier}, 0.75\,s before the saccade; \emph{before} and \emph{after saccade}, the frames on either side of the jump; \emph{before region update}, the last frame before the next SR call; \emph{region update}, the first frame of that call, which moves the sharp region to the new gaze; and \emph{later}, 0.5\,s after that. Ultra Flash super-resolves the whole frame, so its gaze crop is always sharp.}
  \label{fig:supp-timestrip}
\end{figure*}

%% file: sec/X_suppl_experiments.tex
\section{Whole-frame and acuity-filtered KID}
\label{sec:supp-periphery}
Gaze-crop metrics do not see the periphery, so we also compare whole frames with Ultra Flash. Against Ultra Flash from four independent seeds, with the floor (seed 0 against seeds 1--4) subtracted, whole-frame KID ($\times10^{3}$, lower is better; tiny decoder, 1K, 158 prompts) is 13.19 for \textit{Balanced} and 18.94 for \textit{Fast}, against 15.11 for Latent Upsampling and 28.58 for matched-speed Uniform. With the viewing filter applied to both sides, it is 12.53, 17.84, 14.58 and 25.62. \textit{Balanced} and \textit{Fast} thus stay closer to Ultra Flash than Uniform with and without the filter, and Latent Upsampling falls between them.
Gaze-crop KID in the main paper is absolute, and Ultra Flash's own value varies with each job's seed-0 clips (0.63 in Tab.~1 of the main paper, 0.48 in Tab.~4 of the main paper), little against the gaps between methods. Filtering Ultra Flash raises it from 0.63 to 5.10 (paired $+4.47$, 95\% interval $[3.04,6.08]$).
The viewing filter is set at a 1\,JOD loss, the detection threshold at which 75\% of observers prefer one video~\cite{mantiuk2021fovvideovdp,perezortiz2017pwcmp} and to which foveated rendering is commonly calibrated~\cite{patney2016towards,kam2026dichoptic}. Viewers in our study did not distinguish filtered Ultra Flash from Ultra Flash ($+0.02$\,JOD, \cref{sec:supp-study-all}), so the filter is conservative under central fixation.

\section{Component ladder: PSNR and LPIPS}
\label{sec:supp-fidelity}
PSNR and LPIPS against Ultra Flash measure reconstruction only while the LR video is unchanged. Once RA or the KV budget change the generated content, they measure how far the content has moved from Ultra Flash, as in Sec.~4.3 of the main paper, not lost detail.
On $7.5^\circ$ gaze-centered crops of lossless tiny-decoder outputs (158 prompts, every second frame, no acuity filter), foveated SR alone and with foveated decoding, which keep the LR video, stay close to Ultra Flash: 33.50 and 33.45\,dB PSNR, LPIPS 0.035 and 0.037. Adding RA and then Gaze KV changes the content, and the scores fall to 20.21 and 17.94\,dB (LPIPS 0.218 and 0.309).
The same-LR control separates the two effects. It compares \textit{Fast} with full-frame SR and full-resolution decoding from the same RA and KV generator, so both share the LR video. On the 154 prompts whose LR hashes matched in all seven SR groups, \textit{Fast} reaches 35.18\,dB and LPIPS 0.035 (34.78\,dB and 0.042 over all 158 prompts, four of which diverged after the first group). At the gaze, the foveated stages thus stay close to their full-frame counterparts, and the fall against Ultra Flash comes from the changed content.